\documentclass{article}

\PassOptionsToPackage{numbers, compress}{natbib}
\usepackage[preprint]{neurips_2026}

\usepackage{microtype}
\usepackage{graphicx}
\usepackage{subcaption}
\usepackage{booktabs} % for professional tables
\usepackage{multirow}
\usepackage{colortbl} % Required for row color
\definecolor{Gray}{gray}{0.95} % Light gray for "Ours"
\usepackage{arydshln}
\usepackage{placeins}
\usepackage{amsmath}
\usepackage{enumitem}
\usepackage{wrapfig}
\usepackage{hyperref}
\usepackage{graphicx} % Required for \resizebox
\usepackage{booktabs} % Required for \toprule, \midrule, \bottomrule
\usepackage{pifont}   % Required for the check and cross marks
\usepackage{xcolor}

\newcommand{\cmark}{\ding{51}}
\newcommand{\xmark}{\ding{55}}

\usepackage[utf8]{inputenc} % allow utf-8 input
\usepackage[T1]{fontenc}    % use 8-bit T1 fonts
\usepackage{hyperref}       % hyperlinks
\usepackage{url}            % simple URL typesetting
\usepackage{booktabs}       % professional-quality tables
\usepackage{amsfonts}       % blackboard math symbols
\usepackage{nicefrac}       % compact symbols for 1/2, etc.
\usepackage{microtype}      % microtypography
\usepackage{xcolor}         % colors

\title{Hyperbolic Concept Bottleneck Models}
\author{%
  Daniel Uyterlinde \quad Swasti S. Mishra \quad Pascal Mettes \\
  Informatics Institute, University of Amsterdam, The Netherlands \\
  \texttt{daniel.uijterlinde@student.uva.nl, \{s.s.mishra, p.s.m.mettes\}@uva.nl}
}

\begin{document}

\maketitle

\begin{abstract}
  Concept Bottleneck Models (CBMs) have become a popular approach to enable interpretability in neural networks by constraining classifier inputs to a set of human-understandable concepts. While effective, current models embed concepts in flat Euclidean space, treating them as independent, orthogonal dimensions. Concepts, however, are highly structured and organized in semantic hierarchies. To resolve this mismatch, we propose Hyperbolic Concept Bottleneck Models (HypCBM), a post-hoc framework that grounds the bottleneck in this structure by reformulating concept activation as asymmetric geometric containment in hyperbolic space. Rather than treating entailment cones as a pre-training penalty, we show they encode a natural test-time activation signal: the margin of inclusion within a concept's entailment cone yields sparse, hierarchy-aware activations without any additional supervision or learned modules. We further introduce an adaptive scaling law for hierarchically faithful interventions, propagating user corrections coherently through the concept tree. Empirically, HypCBM rivals post-hoc Euclidean models trained on 20$\times$ more data in sparse regimes required for human interpretability, with stronger hierarchical consistency and improved robustness to input corruptions.
\end{abstract}

\section{Introduction}
The advent of large black-box models, particularly vision-language models \cite{radford2021learning, li2022blip, liu2023visual}, across high-stakes domains including healthcare \cite{li2023llava, moor2023med}, finance \cite{huang2024open}, and autonomous systems \cite{tian2024drivevlmconvergenceautonomousdriving}, has intensified the need for models whose predictions are transparent and interpretable. While various frameworks offer post-hoc explanations \cite{hu2023seat, hu2023improving, yang2024human}, Concept Bottleneck Models (CBMs) \cite{koh2020concept} have emerged as a canonical approach by enforcing an intermediate, interpretable bottleneck. By requiring the model to predict human-understandable concepts before making a final decision, this architecture exposes internal reasoning and, crucially, enables users to intervene and correct erroneous predictions at the concept level. CBMs have established themselves in interpretable learning, enabling domain experts to de-bias models and improve reliability through human-in-the-loop interaction \cite{shin2023closer}. 

Although recent frameworks have improved CBM scalability and usability \cite{zarlenga2022conceptembeddingmodelsaccuracyexplainability, oikarinen2023labelfree, steinmann2024learninginterveneconceptbottlenecks}, they retain a limiting representational assumption: concept independence. In real-world settings, concepts are rarely independent, but visually co-occurring and semantically structured, exhibiting relationships ranging from synonymy to inherent hierarchies \cite{barbiero2024relational}. While some approaches enforce structure through auxiliary modules or constraints \cite{pittino2023hierarchical, panousis2024coarse, vemuri2025logiccbmslogicenhancedconceptbasedlearning}, the underlying bottleneck activations are derived from a flat Euclidean embedding space, where concepts are treated as independent, orthogonal directions. This geometry is fundamentally at odds with hierarchical logic: in a tree-like structure, the number of nodes grows exponentially with depth, yet Euclidean space cannot embed branching relationships without forcing parent and child concepts into overlapping or distorted regions \cite{10.1007/978-3-642-25878-7_34}. The result is redundant concept sets, logical inconsistencies (e.g., ``\textit{Dog}'' without ``\textit{Animal}''), and inefficient interventions.

Hyperbolic geometry offers a solution to this mismatch. Its exponential volume growth makes it the natural domain for hierarchical representation learning, a property that has driven advances in taxonomy embedding \cite{nickel2017poincare, ganea2018hyperbolic}, image embeddings \cite{khrulkov2020hyperbolic}, segmentation \cite{atigh2022hyperbolic}, and hierarchical classification \cite{dhall2020hierarchicalimageclassificationusing, 9157196}. Recently, hyperbolic vision-language models \cite{desai2023hyperbolic, pal2024compositional} have shown that this geometry can be aligned with semantic structure at scale: through entailment cone training objectives, parent concepts come to geometrically subsume their children. We show that this geometric structure encodes a natural concept activation signal that is hierarchically consistent by construction and directly interpretable at inference time, without any additional supervision or learned modules. 

Building on this insight, we introduce Hyperbolic Concept Bottleneck Models (HypCBM). Rather than relying on symmetric distance metrics that distort hierarchical relationships, we reformulate concept activation as an asymmetric geometric containment problem in hyperbolic space. Specifically, our contributions are:
\begin{itemize}[leftmargin=*,itemsep=0.5pt, topsep=0pt]
    \item We propose a \textbf{hyperbolic entailment activation mechanism} that maps image representations to sparse, hierarchy-aware concept activations through entailment cone containment.

    \item We introduce an \textbf{adaptive scaling law for hierarchical interventions} to propagate corrections coherently through the concept tree. Because hyperbolic volume expands exponentially with depth, a fixed entailment threshold fails; we formulate a scaling mechanism that adapts strictness to the geometric specificity of the parent concept.

    \item We demonstrate that HypCBM consistently \textbf{improves accuracy} over existing post-hoc CBMs, rivaling Euclidean models trained on 20$\times$ more data in sparse regimes, while also enhancing data efficiency, robustness, and hierarchical consistency. 
\end{itemize}

% We operationalize interpretability in CBMs in this work through four properties: 
% (i) sparsity, ensuring concise explanations; 
% (ii) hierarchical consistency, enforcing semantic coherence; 
% (iii) intervention faithfulness, enabling predictable human corrections; and 
% (iv) stability, requiring explanations to remain invariant under benign input perturbations.

\section{Related Work}
\subsection{Concept Bottleneck Models}
%\daniel{moved to related work}
Concept Bottleneck Models (CBMs) represent a prominent paradigm in explainable machine learning. Unlike classical attribution methods (e.g., LIME \cite{ribeiro2016whyitrustyou}, SHAP \cite{lundberg2017unifiedapproachinterpretingmodel}), CBMs are interpretable-by-design. First proposed by \citet{koh2020concept}, CBMs align the predictive process of neural networks with human reasoning by introducing an intermediate layer of semantically meaningful concepts. This is done by decomposing the classification task into two mappings: a concept extractor $f:\mathcal{X} \rightarrow \mathbb{R}^K$ that maps raw inputs $\mathbf{x}$ to a concept activation vector $\mathbf{c} \in \mathbb{R}^K$, and an interpretable predictor $g:\mathbb{R}^K \rightarrow \mathcal{Y}$ predicting label $\mathbf{y}$ based solely on $\mathbf{c}$. This bottleneck crucially enables test-time intervention: a human expert can correct wrongly predicted concept activations to correct the final model prediction \cite{shin2023closer}.
\begin{table}[b]
\centering
\vspace{-2em}
\caption{\textbf{Taxonomy of popular CBM methods.} HypCBM uniquely achieves a structured bottleneck completely post-hoc without learned modules. A full overview is provided in Appendix~\ref{app:taxonomy}.}
\label{tab:cbm_taxonomy}
\resizebox{\textwidth}{!}{
\begin{tabular}{lcccl}
\toprule
\textbf{Method} & \textbf{Post-hoc} & \textbf{Zero-shot Concepts} & \textbf{Structured Bottleneck} & \textbf{Learned Modules}\\
\midrule
LaBo \cite{yang2023languagebottlelanguagemodel} & \cmark & \cmark & \xmark & Optimized concept selection  \\
VLG-CBM \cite{srivastava2025vlgcbmtrainingconceptbottleneck} & \xmark & \xmark & \xmark & Grounding + learned projection \\
Hierarchical CBM \cite{pittino2023hierarchical} & \xmark & \xmark & \cmark & Supervised hierarchy layers \\
CRM \cite{debot2025interpretablehierarchicalconceptreasoning} & \xmark & \xmark & \cmark & Graph reasoning  \\
LogicCBM \cite{vemuri2025logiccbmslogicenhancedconceptbasedlearning} & \xmark & \xmark & \cmark & Neuro-symbolic reasoning \\
PCBM \cite{yuksekgonul2023posthoc} / LF-CBM \cite{oikarinen2023labelfree} & \cmark & \cmark & \xmark & \textbf{None}  \\
\midrule
\textbf{HypCBM (Ours)} & \cmark & \cmark & \cmark & \textbf{None} \\
\bottomrule
\end{tabular}}
\end{table}

Despite their promise, standard CBMs require dense concept supervision, limiting scalability. Early post-hoc interpretation methods such as TCAV \cite{kim2018interpretabilityfeatureattributionquantitative} mitigated this by training linear classifiers per concept, but still required per-concept annotation. Post-hoc CBMs (PCBMs) \cite{yuksekgonul2023posthoc} and their label-free variant LF-CBM \cite{oikarinen2023labelfree} eliminated this by exploiting the shared embedding space of pre-trained VLMs: concept vectors $\mathbf{v}_i$ are retrieved via the text encoder, and activation is computed as cosine similarity $c_i = \langle f(\mathbf{x}), \mathbf{v}_i \rangle$, converting any backbone into an interpretable model without additional training. Subsequent frameworks extended this through language-supervised bottlenecks \cite{yang2023languagebottlelanguagemodel}, learned concept embeddings \cite{zarlenga2022conceptembeddingmodelsaccuracyexplainability}, unsupervised concept discovery \cite{rao2024discoverthennametaskagnosticconceptbottlenecks}, and vision-grounded concept annotation \cite{srivastava2025vlgcbmtrainingconceptbottleneck}.

A separate line of work addresses the structural limitation of flat concept spaces by enforcing explicit semantic relationships. These approaches achieve structural validity through supervised hierarchy layers \cite{pittino2023hierarchical}, probabilistic coarse-to-fine modeling \cite{panousis2024coarse}, or neuro-symbolic rules and graph-based reasoning \cite{vemuri2025logiccbmslogicenhancedconceptbasedlearning, defelice2025causallyreliableconceptbottleneck, debot2025interpretablehierarchicalconceptreasoning}. However, all such methods require learned modules and depart from the post-hoc paradigm, entangling the effect of structural inductive bias with that of additional supervision. \textbf{Direct comparison with these methods is therefore neither fair nor feasible:} they require task-specific training incompatible with our post-hoc setting, and their auxiliary modules are designed for flat Euclidean spaces and cannot be natively extended to hyperbolic geometry.
\begin{figure}[t]
    \centering
    \includegraphics[width=\linewidth]{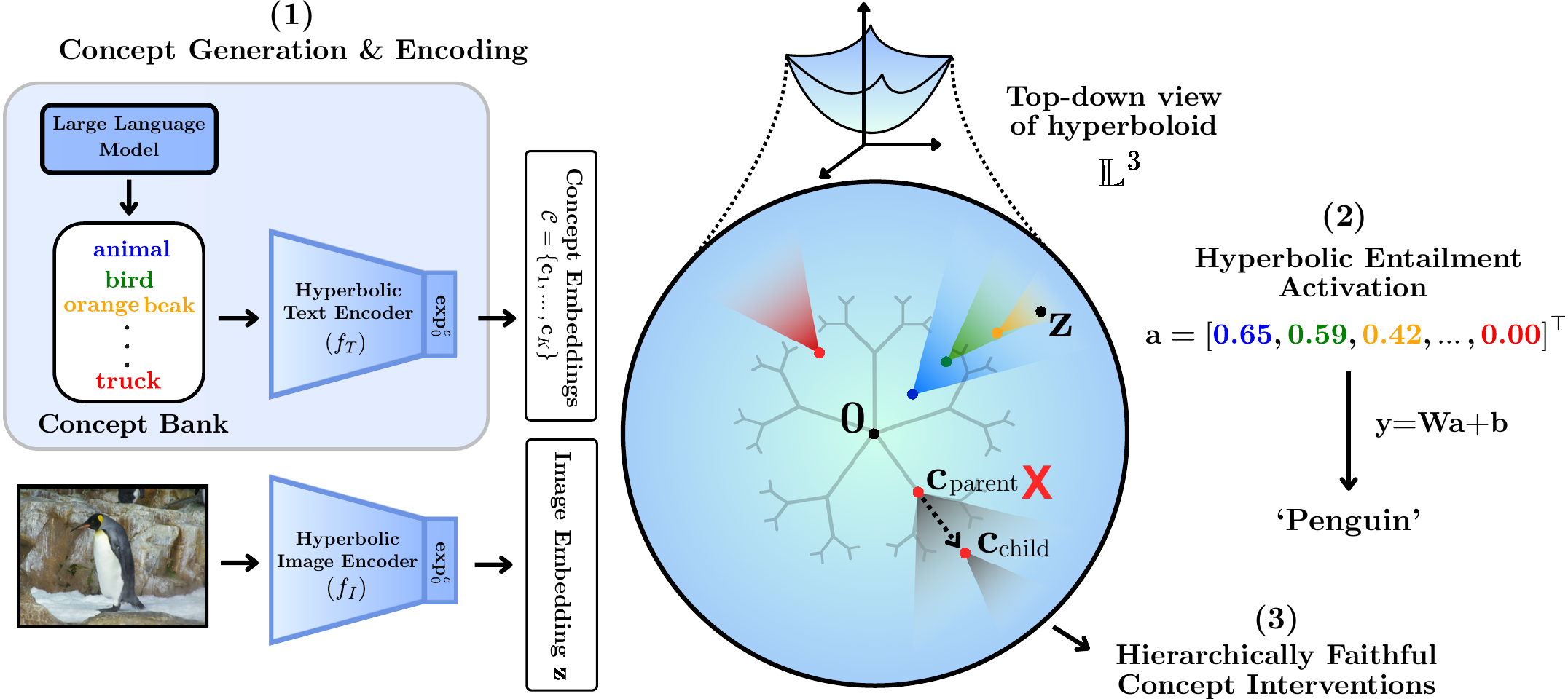}
    \caption{\textbf{Method overview.} \textbf{(1)} Generated concepts and the target image are encoded with a hyperbolic VLM onto the hyperbolic manifold, where they are hierarchically organized. \textbf{(2)} The activation of a concept is measured as the \textit{margin of inclusion} of the image embedding in the entailment cone of the concept. \textbf{(3)} An intervention on a parent concept $(\mathbf{c}_{\text{parent}})$ is propagated to all entailed children $(\mathbf{c}_{\text{child}})$.}
    \label{fig:method}
    \vspace{-1.4em}
\end{figure}
\subsection{Hyperbolic Vision-Language Models}
Standard vision-language models like CLIP \cite{radford2021learning} and ALIGN \cite{jia2021scaling} align image and text encoders via a symmetric contrastive objective. While remarkably effective for zero-shot generalization \cite{ramesh2022hierarchicaltextconditionalimagegeneration, Rombach_2022_CVPR}, their reliance on Euclidean geometry limits their ability to capture the intrinsic hierarchy of visual concepts \cite{desai2023hyperbolic}. Recent works address this by generalizing VLMs to hyperbolic space, whose exponential volume growth naturally accommodates tree-like semantic structures. MERU \cite{desai2023hyperbolic} pioneered this direction by redefining image-text alignment as an entailment problem, using hyperbolic entailment cones to enforce asymmetric containment. HyCoCLIP \cite{pal2024compositional} extended this to compositional object-part hierarchies, and HySAC \cite{poppi2025hyperbolicsafetyawarevisionlanguagemodels} applied the same principle for safety-aware representations. We leverage these pre-trained hyperbolic backbones as the geometric foundation for constructing hierarchically faithful concept bottlenecks. Given the rapid growth of entailment-based hyperbolic VLMs \cite{Ramasinghe_2024_CVPR, huynh2026argentadaptivehierarchicalimagetext}, HypCBM presents a zero-shot pathway to interpretability for this emerging model class, with concept hierarchy emerging naturally from the underlying geometry.

% As the landscape of entailment-based hyperbolic VLMs rapidly expands \cite{Ramasinghe_2024_CVPR, huynh2026argentadaptivehierarchicalimagetext}, HypCBM provides a zero-shot mechanism to turn this emerging class of models into interpretable classifiers. 

Table~\ref{tab:cbm_taxonomy} positions HypCBM within this landscape. It is the only framework that achieves a structured bottleneck while remaining fully post-hoc with no learned modules. By treating the backbone as a frozen feature extractor, any empirical gains over post-hoc Euclidean baselines are attributable directly to the geometric formulation, free from confounding supervision artifacts.

\section{Methodology}
\label{sec:method}
We introduce HypCBM, a post-hoc framework that enriches concept bottlenecks with a hierarchical organization through hyperbolic geometry. We begin by formalizing the necessary preliminaries of hyperbolic space. We then detail the concept generation and encoding phase, followed by our two core technical contributions: a hyperbolic entailment activation mechanism and an adaptive strategy for hierarchically faithful concept interventions.
% \subsection{Preliminaries on Hyperbolic Geometry}
% We encode concepts and images in a hyperbolic embedding space: a Riemannian manifold of constant negative curvature $-c$, $c>0$. We utilize the Lorentz model (hyperboloid model), known for its superior numerical stability over the Poincaré ball \cite{nickel2018learning, numericstability-mishne2023}. The $n$-dimensional Lorentz model $\mathbb{L}^n_c$ is defined as:
% \begin{equation}
%     \mathbb{L}^n= \left\{ \mathbf{x} \in \mathbb{R}^{n+1} : 
%     \langle \mathbf{x}, \mathbf{x} \rangle_{\mathbb{L}} = 
%     -\frac{1}{c},\ c > 0 \right\}.
% \end{equation}
% A point $\mathbf{x}\in\mathbb{L}^n_c$ decomposes into a time component 
% $x_0$ and spatial component $\mathbf{\tilde{x}} \in \mathbb{R}^n$, where 
% $x_0 = \sqrt{1/c + \|\mathbf{\tilde{x}}\|^2}$ is fully determined by 
% $\mathbf{\tilde{x}}$. This allows us to work exclusively with the 
% $n$-dimensional spatial component while respecting manifold geometry. 
% Critically, this space encodes hierarchy via \textit{radial ordering}: 
% generic concepts cluster near the origin, while specific concepts are 
% pushed toward the boundary \cite{pal2024compositional}; the geometric 
% structure our activation mechanism exploits directly.

\subsection{Preliminaries on Hyperbolic Geometry}
We encode concepts and images in a hyperbolic embedding space, a Riemannian manifold of constant negative curvature $-c$, with $c>0$. While several isometric models of hyperbolic space exist (e.g., the Poincaré ball), we utilize the Lorentz model (or hyperboloid model), known for its superior numerical stability for optimization compared to the Poincaré ball, particularly avoiding vanishing gradients near the boundary, and has become a popular choice for hyperbolic representation learning \cite{nickel2018learning, numericstability-mishne2023}.

Formally, the $n$-dimensional Lorentz model $\mathbb{L}^n_c$ is defined as the upper sheet of a two-sheeted hyperboloid in an ($n+1$)-dimensional Minkowski spacetime \cite{cannon1997hyperbolic}:
\begin{equation}
    \mathbb{L}^n_c= \left\{ \mathbf{x} \in \mathbb{R}^{n+1} : \langle \mathbf{x}, \mathbf{x} \rangle_{\mathbb{L}} = -\frac{1}{c}, c > 0 \right\}.
\end{equation}
% A point $\mathbf{x}\in\mathbb{L}^n_c$ can be decomposed into a \textit{time} component $x_0$ and a \textit{space} component $\mathbf{x}_{1:n} \in \mathbb{R}^n$, which we denote as $\mathbf{\tilde{x}}$ for brevity. Crucially, the manifold constraint $\langle \mathbf{x}, \mathbf{x} \rangle_{\mathbb{L}}=-1/c$ implies that the temporal coordinate is a dependent variable uniquely determined by the spatial coordinates: 
% \begin{equation} 
% x_0 = \sqrt{1/c + \|\mathbf{\tilde{x}}\|^2}. 
% \end{equation} This allows us to represent hyperbolic embeddings using only their $n$-dimensional spatial components while implicitly respecting the manifold geometry.
A point $\mathbf{x}\in\mathbb{L}^n_c$ decomposes into a time component 
$x_0$ and spatial component $\mathbf{\tilde{x}} \in \mathbb{R}^n$, where 
$x_0 = \sqrt{1/c + \|\mathbf{\tilde{x}}\|^2}$ is fully determined by 
$\mathbf{\tilde{x}}$. This allows us to work exclusively with the 
$n$-dimensional spatial component while respecting manifold geometry. 

Standard neural networks (like the CLIP image and text encoders) produce features $\mathbf{v}\in \mathbb{R}^n$ in a flat Euclidean space. To map these features onto the hyperbolic manifold, we utilize the exponential map at the origin, $\mathbf{o} = (\sqrt{1/c}, \mathbf{0})^\top$. This maps a tangent vector $\mathbf{v}$ (identified with $\mathbb{R}^n$) to $\mathbb{L}^n_c$ via:
\begin{equation}
\text{exp}_{\mathbf{0}}(\mathbf{v}) = \cosh(\sqrt{c} \|\mathbf{v}\|) \mathbf{0} + \frac{\sinh(\sqrt{c} \|\mathbf{v}\|)}{\sqrt{c} \|\mathbf{v}\|} \mathbf{v}.
\end{equation}
Inverse operations map points back to the tangent space via the logarithmic map $\text{log}_\mathbf{0}(\cdot)$, enabling standard Euclidean operations when necessary.

\subsection{Concept Generation and Encoding}
\label{sec:conceptgeneration}
Following the label-free, post-hoc paradigm of \citet{oikarinen2023labelfree}, we automate concept discovery without manual annotations in two steps: generating a candidate set of visually descriptive attributes and encoding them onto the hyperbolic manifold.

For concept generation, we prompt an LLM with dataset-specific class labels to produce visually descriptive attributes per class. These per-class lists are aggregated into a global set of unique textual descriptions $\mathcal{T}=\{t_1,\dots,t_M\}$. Full prompts and generation examples are provided in Appendix~\ref{app:concept bank}.

The resulting descriptions are embedded into hyperbolic space using the frozen text encoder $f_T: \mathcal{T} \to \mathbb{L}^n_c$ of any pre-trained hyperbolic VLM. The initial concept bank $\mathcal{C}_{\text{init}}$ of size $M$ is defined as:
% Our framework adopts the label-free, post-hoc paradigm of \citet{oikarinen2023labelfree}, which automates concept discovery without requiring manual annotations. This process consists of two steps: generating a candidate set of visually descriptive attributes and encoding these descriptions into the hyperbolic manifold. 
% For the generation phase, we prompt GPT-3 with dataset-specific class labels to produce visually descriptive attributes. These per-class lists are then aggregated into a global, unique set of textual concept descriptions, $\mathcal{T}=\{t_1,\dots,t_M\}$. Full prompts and generation examples are provided in Appendix~\ref{app:concept bank}.
% To embed textual descriptions in hyperbolic space, we utilize the pre-trained, frozen encoders of any hyperbolic vision-language model.
% Let $f_T: \mathcal{T} \to \mathbb{L}^n_c$ denote the text encoder. We define the initial concept bank $\mathcal{C}_{\text{init}}$ of size $M$ by encoding the global vocabulary $\mathcal{T}$:
\begin{equation}
    \mathcal{C}_{\text{init}} = \left\{ \mathbf{c}_i \in \mathbb{L}^n_c \mid \mathbf{c}_i = f_T(t_i) \right\}_{i=1}^{M}.
\end{equation}
% Crucially, this embedding space encodes hierarchy via radial ordering: semantically generic concepts cluster near the origin, while specific concepts are pushed toward the boundary \cite{pal2024compositional}. We exploit this intrinsic organization to extract hierarchically faithful concept activations.
Crucially, each vector $\mathbf{c}_i$ is positioned according to the semantic hierarchy learned by the backbone during pre-training: specific concepts lie within the cones of their generic parents, with correspondingly larger spatial norms \cite{desai2023hyperbolic, pal2024compositional}. We exploit this intrinsic organization in the activation mechanism described next.
% Crucially, each vector $\mathbf{c}_i$ is positioned according to the semantic hierarchy learned during pre-training: generic concepts map near the origin, while specific attributes are pushed toward the boundary with larger spatial norms due to specific instances being entailed by general ones. We exploit this intrinsic organization in the activation mechanism described next.

\paragraph{Norm-Based Filtering.} While generated concept banks provide rich features, they often contain generic terms (e.g., `object', `entity') that lack discriminative power. By embedding concepts as vectors in a hyperbolic embedding space, we are able to filter out overly generic concepts simply by computing their distance to the origin. Overly generic concepts inflate the size of the concept bank without adding significant predictive value since such concepts are applicable to many samples.
We therefore apply a geometric filter to prune them, defining the final concept bank $\mathcal{C}$ as: 
\begin{equation}
    \mathcal{C} = \{ \mathbf{c} \in \mathcal{C}_{init} \mid \|\tilde{\mathbf{c}}\| \ge \tau \}.
\end{equation}
We set $\tau=0.27$ based on the HyCoCLIP \cite{pal2024compositional} concept embeddings, filtering the bottom $\approx20^{\text{th}}$ percentile of non-discriminative root concepts. To ensure a fair comparison, we utilize this resulting concept bank for both the Euclidean and hyperbolic models. We empirically validate in Appendix~\ref{sec:ablation} that this geometric filtering is critical for hyperbolic models, whereas Euclidean baselines show marginal gains from similar pruning. The final concept counts ($N_c$) per dataset are reported in Table~\ref{tab:main_results}.
\begin{figure*}[t]
    \centering
    % Panel A: Entailment Cone
    \begin{subfigure}[c]{0.45\textwidth}
        \centering
        \includegraphics[width=\linewidth]{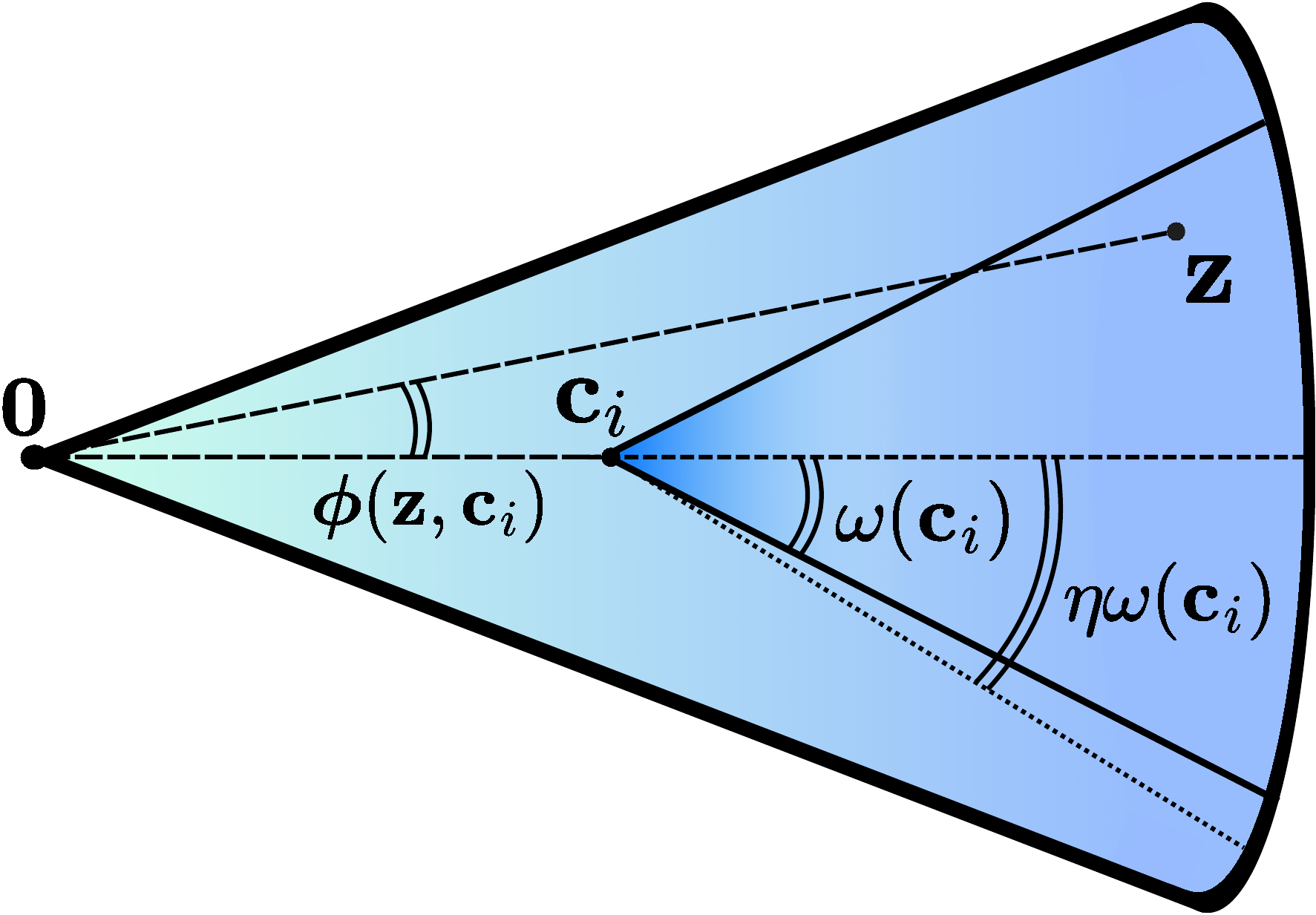}
        \caption{Geometric Entailment}
        \label{fig:entailmentcone}
    \end{subfigure}\hfill
    % Panel B: Scaling Law
    \begin{subfigure}[c]{0.5\textwidth}
        \centering
        \includegraphics[width=\linewidth]{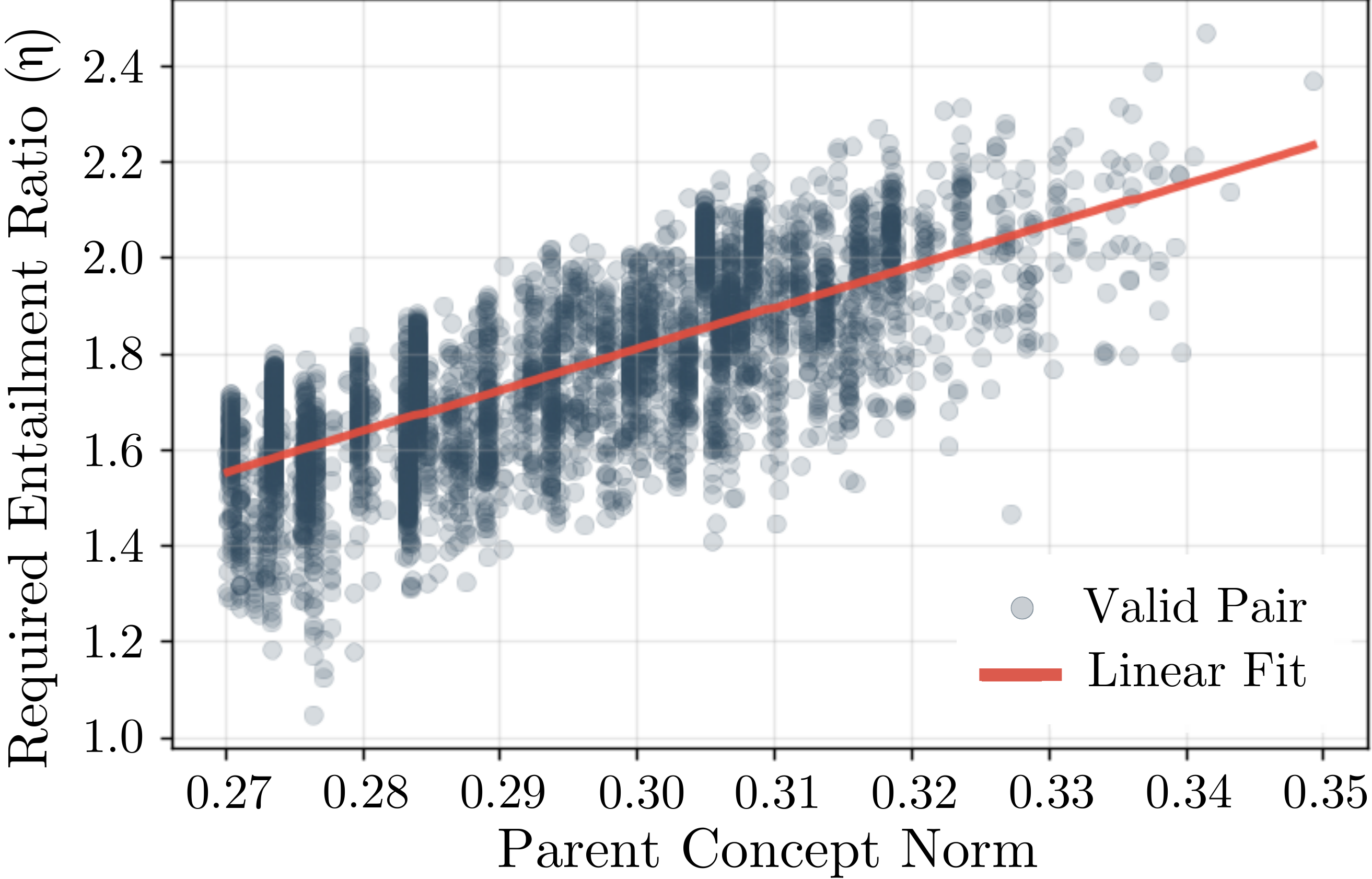}
        \caption{Concept-to-Concept Scaling Law}
        \label{fig:scaling_law}
    \end{subfigure}
    \caption{\textbf{Geometry and Scaling of Hyperbolic Entailment.} \textbf{(a)} An image $\mathbf{z}$ activates concept $\mathbf{c}_i$ if the exterior angle $\phi(\mathbf{z}, \mathbf{c}_i)$ falls within the scaled cone half-aperture $\eta \omega(\mathbf{c}_i)$. Here, $\eta$ denotes the strictness parameter $\eta_{\text{img}}$ (Eq.~\ref{eq:activation}). \textbf{(b)} Empirical scaling law derived from WordNet. The entailment strictness $\eta_{\text{text}}$ required to geometrically capture true descendants scales linearly with the parent concept's norm ($r=0.729$).}
    \vspace{-1em}
    \label{fig:geometry_and_scaling}
\end{figure*}

\subsection{Hyperbolic Entailment Activation}
\label{sec:activation}
Unlike standard symmetric distance metrics (e.g., Euclidean cosine similarity), HypCBM formulates concept activation as an asymmetric geometric containment problem on the Lorentz manifold. While prior work restricts entailment cones to pre-training penalties \cite{ganea2018hyperbolic, desai2023hyperbolic, pal2024compositional}, we show that the resulting geometric structure serves directly as an inference mechanism. 

Specifically, the concept activation $a_i$ measures the degree to which image embedding $\mathbf{z}$ is geometrically contained within the entailment cone of concept $\mathbf{c}_i$ (see Fig.~\ref{fig:entailmentcone}), yielding a sparse, geometrically grounded activation vector that is hierarchically consistent by construction. 

The aperture of this cone is determined by the concept's specificity. Following 
\citep{le2019inferringconcepthierarchiestext, desai2023hyperbolic}, we define the cone's 
half-aperture $\omega(\mathbf{c}_i)$ as inversely proportional to the concept's spatial norm, with scaling constant $K=0.04$ balancing cone expressivity against saturation 
(see Appendix~\ref{app:distribution}):
\begin{equation}
    \omega(\mathbf{c}_i) = \sin^{-1} \left( \frac{2K}{\sqrt{c} \| \tilde{\mathbf{c}}_i \|} \right).
\end{equation}
To determine whether the image embedding $\mathbf{z}$ falls within the entailment cone, we compute the \textit{exterior angle} $\phi(\mathbf{z}, \mathbf{c}_i)$ relative to the concept axis \cite{desai2023hyperbolic, pal2024compositional}:
\begin{equation}
    \phi(\mathbf{z}, \mathbf{c}_i) = \cos^{-1}\left(\frac{z_0 + (\mathbf{c}_i)_0 \, c \langle \mathbf{z}, \mathbf{c}_i \rangle_{\mathbb{L}}}{\|\tilde{\mathbf{c}}_i\| \sqrt{(c \langle \mathbf{z}, \mathbf{c}_i \rangle_{\mathbb{L}})^2 - 1}}\right).
    \label{eq:exterior}
\end{equation}
Geometrically, the partial order relation is satisfied when this exterior angle is strictly smaller than the cone's half-aperture. We define the concept activation as the \textit{margin of inclusion}, normalized by the cone aperture:
\begin{equation} 
    a_{i} = \max\left(0, \eta_{\text{img}} - \frac{\phi(\mathbf{z}, \mathbf{c}_i)}{\omega(\mathbf{c}_i)}\right),
    \label{eq:activation}
\end{equation}
resulting in a naturally sparse, non-negative activation vector $\mathbf{a}=[a_1,\dots,a_{M}]^\top$ where $a_i > 0$ implies geometric entailment. Here, $\eta_{\text{img}}$ is a strictness parameter calibrated to separate true positive entailments from false associations (see Appendix~\ref{app:eta_img}). Normalizing by $\omega(\mathbf{c}_i)$ measures \emph{relative containment depth}, or the ratio of remaining margin to total cone aperture, making activations comparable across concepts of varying specificities: an identical activation value indicates the same relative geometric alignment regardless of whether the concept is generic or specific. Overly generic concepts with broad cones that would activate across most images are further suppressed by norm-based filtering (Section~\ref{sec:conceptgeneration}).

% Because general concepts have wider cones, they naturally yield higher absolute activations for a given angular distance, geometrically mirroring the logical certainty of semantic classification. 

Finally, we predict class logits using a sparse linear layer $g(\mathbf{a}) = \mathbf{W}\mathbf{a} + \mathbf{b}$, where $\mathbf{W}$ is regularized with Elastic-Net ($\alpha=0.99$) to ensure decisions rely on a small set of active concepts, with sparsity controlled by regularization parameter $\lambda$.

\subsection{Hierarchically Faithful Concept Interventions}
\label{sec:method interventions}
A critical limitation of Euclidean CBMs is the lack of structural consistency during test-time interventions. Manually deactivating a parent concept (e.g., \textit{``Animal''}) leaves its child concepts (e.g., \textit{``Dog''}) active, resulting in logical paradoxes. HypCBM resolves this by automatically propagating interventions through the semantic concept hierarchy: when a user suppresses a parent concept, all geometric descendants are as well (see Fig.~\ref{fig:method}).

A concept $\mathbf{c}_{\text{child}}$ is identified as a descendant of $\mathbf{c}_{\text{parent}}$ if it falls within the parent's entailment cone under a text-to-text strictness threshold $\eta_{\text{text}}$. However, a fixed threshold is insufficient across hierarchy depths: due to the exponential volume growth of hyperbolic space, the entailment margin required to correctly capture true descendants varies with the parent's specificity.

To address this, we derive an adaptive scaling law for $\eta_{\text{text}}$ from 10k WordNet \citep{10.1145/219717.219748} pairs, finding that the required entailment margin scales linearly with the parent's norm (Fig.~\ref{fig:scaling_law},  Appendix~\ref{app:eta_text}). When a user suppresses a concept $\mathbf{c}_{\text{parent}}$ (i.e., setting $a_{\text{parent}} \to 0$), we automatically suppress all $\mathbf{c}_{\text{child}}$ concepts satisfying: 
\begin{equation}
    \phi(\mathbf{c}_{\text{child}}, \mathbf{c}_{\text{parent}}) < \eta_{\text{text}}(\|\mathbf{\tilde{c}}_{\text{parent}}\|) \cdot \omega(\mathbf{\tilde{c}}_{\text{parent}}).
\end{equation}
This allows users to prune entire branches of spurious concepts with a single interaction, substantially reducing the number of interventions required to correct a prediction.

% \begin{figure}[h]
%     \centering
%     \includegraphics[width=0.5\linewidth]{figures/interventionexample.pdf}
%     \caption{\textbf{Hierarchical propagation of interventions:} HypCBM (right) propagates the deactivation of \textit{'commercial space}' to its entailed children ('\textit{venue}', '\textit{public space}'), correcting the prediction to '\textit{apartment building}'. The Euclidean baseline (left) leaves child concepts active and the prediction incorrect.}
%     \label{fig:placeholder}
% \end{figure}
\begin{figure*}[t]
    \centering
    % Left Panel: Global Top Activations
    \begin{subfigure}[b]{0.49\linewidth}
        \centering
        \includegraphics[width=\linewidth]{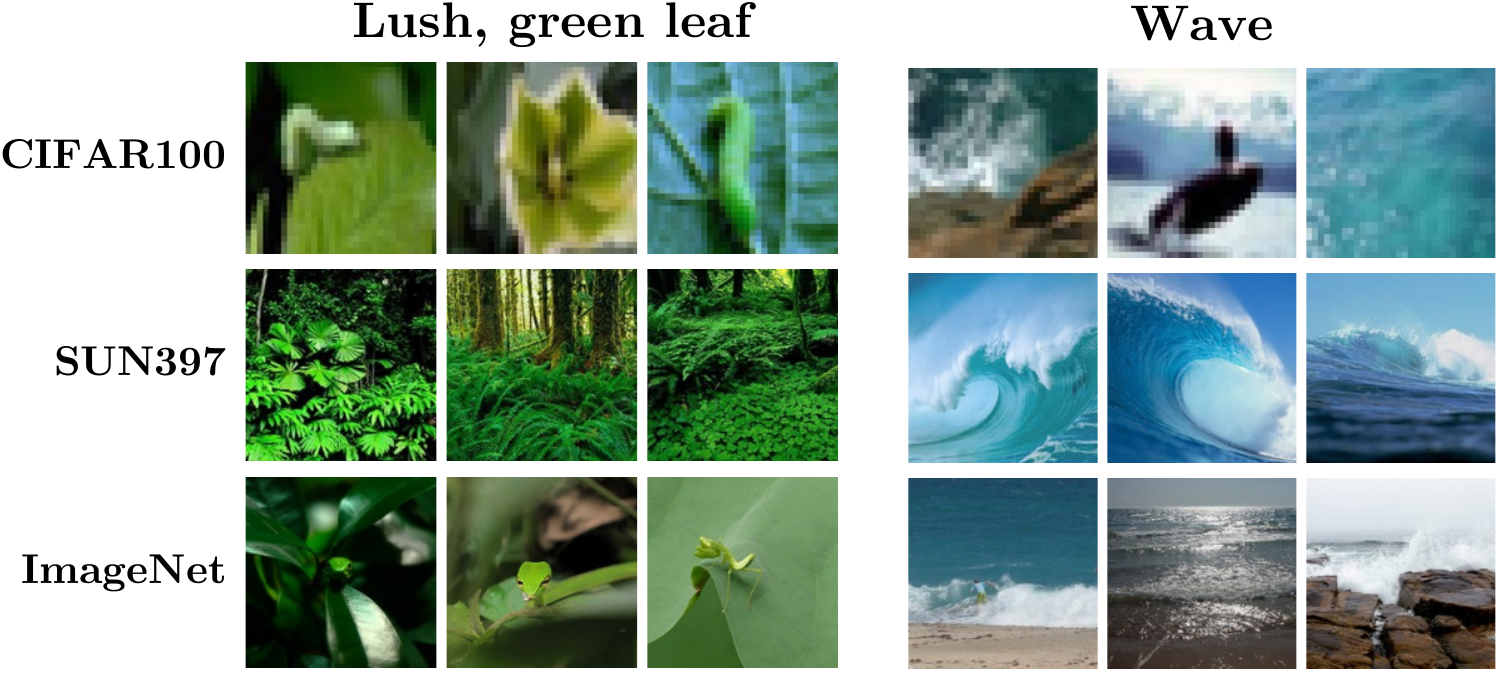}
        \caption{\textbf{Top activating images.}}
        \label{fig:top_activations}
    \end{subfigure}
    \hfill % Pushes them apart
    % Right Panel: Local Explanation
    \begin{subfigure}[b]{0.48\linewidth}
        \centering
        \includegraphics[width=\linewidth]{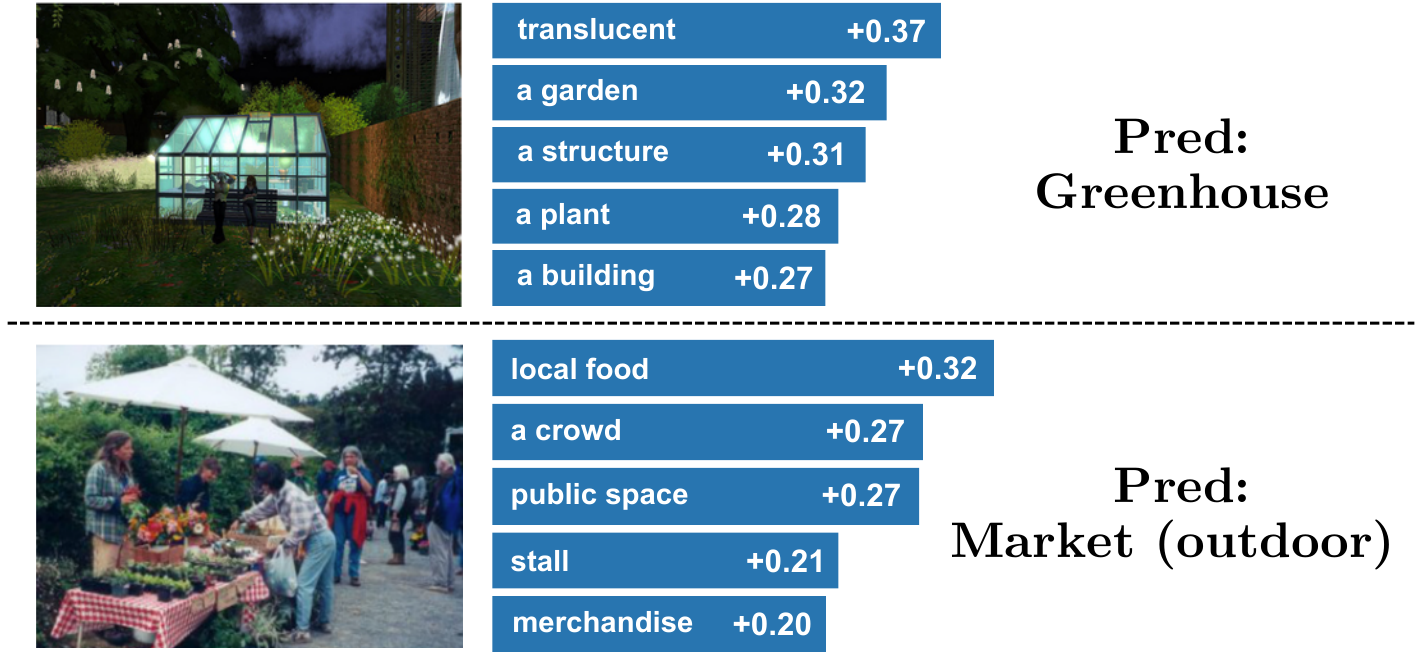}
        \caption{\textbf{Explaining Decisions.}}
        \label{fig:local_explanation}
    \end{subfigure}
    \vspace{-0.2em}
    \caption{We validate the interpretability of HypCBM through \textbf{(a)} inspection of the top activating images per concept to verify semantic alignment, and \textbf{(b)} decision explanations showing the top activating concepts for two images, from ImageNet (top) and SUN397 (bottom).}
    \label{fig:qualitative}
    \vspace{-1em}
\end{figure*}
\section{Experiments}
\vspace{-0.1em}
% \paragraph{Datasets.}
We evaluate HypCBM across three domains: \textbf{CIFAR-100} \cite{Krizhevsky09learningmultiple} for general object classification, \textbf{SUN397} \cite{5539970} for (hierarchical) scene understanding, and \textbf{ImageNet} \cite{5206848} to assess scalability to real-world complexity. Additional results on \textbf{CUB-200} \cite{WahCUB_200_2011} are provided in Appendix~\ref{additional results}. To isolate the impact of geometry, we control for training data by utilizing the pre-trained checkpoints from \citet{pal2024compositional}, ensuring both the hyperbolic HyCoCLIP and Euclidean CLIP \cite{radford2021learning} are trained on the same GrIT-20M dataset \cite{peng2023kosmos2groundingmultimodallarge}. We also report results for CLIP \cite{radford2021learning}, trained on 400M samples, to contextualize performance against a high-resource upper bound.
\begin{table*}[t]
    \centering
    \caption{\textbf{Accuracy at fixed concept budgets.} Test accuracy at strict concept budgets ($K$). HypCBM consistently outperforms Euclidean baselines (LF-CBM) on identical backbones (HyCoCLIP-20M) and often even baselines that use large-scale pre-trained backbones (CLIP-400M and CLIP$^{\text{RE}}$-400M). Linear probing accuracy is reported as an upper bound reference.}
    \vspace{-0.5em}
    \resizebox{\linewidth}{!}{%
    \begin{tabular}{lllccccccc}
    \toprule
       & & & \multicolumn{6}{c}{\textbf{Accuracy @ Concept Budget (ANEC-$K$)}} & \\
       \cmidrule(lr){4-9}
       \textbf{Dataset} & \textbf{Method} & \textbf{Backbone} & \textbf{5} & \textbf{10} & \textbf{20} & \textbf{50} & \textbf{100} & \textbf{Peak} & \textbf{LP}  \\
       \midrule
       \multirow{5}{*}{\shortstack[c]{\textbf{CIFAR-100}\\$(N_c=613)$}} 
         & LF-CBM & CLIP-20M & 39.7 & 48.9 & 55.4 & 60.6 & 64.5 & 67.9 & 73.1 \\
         & LF-CBM & HyCoCLIP-20M  & 25.6 & 34.6 & 56.3 & 61.6 & 62.6 & 69.5 & 76.3 \\
         & \cellcolor{Gray}\textbf{HypCBM (Ours)} & \cellcolor{Gray}HyCoCLIP-20M  & \cellcolor{Gray}\textbf{44.3} & \cellcolor{Gray}\textbf{54.4} & \cellcolor{Gray}\textbf{60.9} & \cellcolor{Gray}\textbf{69.7} & \cellcolor{Gray}\textbf{71.8} & \cellcolor{Gray}\textbf{72.2} & \cellcolor{Gray}76.3 \\
         \cmidrule{2-10}
         & \textit{Ref: LF-CBM} & \textit{CLIP-400M}  & \textit{28.0} & \textit{\underline{58.0}} & \textit{\underline{63.9}} & \textit{64.7} & \textit{67.0} & \textit{71.3} & \textit{82.6} \\
         % & \textit{Ref: LF-CBM} & \textit{CLIP$^{RE}$-400M} & \textit{20.4} & \textit{26.0} & \textit{31.7} & \textit{46.6} & \textit{58.3} & \textit{70.2} & \textit{80.3} \\
       \midrule
       \multirow{5}{*}{\shortstack[c]{\textbf{SUN397}\\$(N_c=1935)$}} 
         & LF-CBM & CLIP-20M         & 45.8 & 53.4 & 61.3 & 67.0 & 68.6 & 70.5 & 74.8 \\
         & LF-CBM & HyCoCLIP-20M   & 34.3 & 44.2 & 50.4 & 59.1 & 61.5 & 72.1 & 75.9 \\
         & \cellcolor{Gray}\textbf{HypCBM (Ours)} & \cellcolor{Gray}HyCoCLIP-20M   & \cellcolor{Gray}\textbf{49.3} & \cellcolor{Gray}\textbf{58.3} & \cellcolor{Gray}\textbf{65.7} & \cellcolor{Gray}\textbf{69.9} & \cellcolor{Gray}\textbf{71.3} & \cellcolor{Gray}\textbf{72.3} & \cellcolor{Gray}75.9 \\
         \cmidrule{2-10}
         & \textit{Ref: LF-CBM} & \textit{CLIP-400M} & \textit{40.0} & \textit{48.0} & \textit{58.8} & \textit{64.4} & \textit{66.8} & \textit{\underline{79.0}} & \textit{82.7} \\
         % & \textit{Ref: LF-CBM} & \textit{CLIP$^{RE}$-400M} & \textit{26.9} & \textit{34.4} & \textit{40.1} & \textit{51.5} & \textit{59.9} & \textit{73.2} & \textit{80.5} \\
       \midrule
       \multirow{4}{*}{\shortstack[c]{\textbf{ImageNet}\\$(N_c=2950)$}} 
         & LF-CBM & CLIP-20M  & 7.7 & 15.3 & 30.7 & 47.3 & 53.7 & 55.3 & 63.7 \\
         & LF-CBM & HyCoCLIP-20M     & 10.2 & 15.5 & 26.0 & 40.6 & 46.2 & 57.9 & 65.4 \\
         & \cellcolor{Gray}\textbf{HypCBM (Ours)} & \cellcolor{Gray}HyCoCLIP-20M  & \cellcolor{Gray}\textbf{17.9} & \cellcolor{Gray}\textbf{34.6} & \cellcolor{Gray}\textbf{44.2} & \cellcolor{Gray}\textbf{52.3} & \cellcolor{Gray}\textbf{56.8} & \cellcolor{Gray}\textbf{58.1} & \cellcolor{Gray}65.4 \\
         \cmidrule{2-10}
         & \textit{Ref: LF-CBM} & \textit{CLIP-400M}  & \textit{17.4} & \textit{34.0} & \underline{\textit{51.8}} & \underline{\textit{52.6}} & \textit{54.0} & \textit{\underline{73.8}} & \textit{78.5} \\
         % & \textit{Ref: LF-CBM} & \textit{CLIP$^{RE}$-400M} &  &  &  &  &  &  & \\
       \bottomrule
    \end{tabular}
    }
    \label{tab:main_results}
    \vspace{-0.5em}
\end{table*}
For all comparisons involving activation sparsity (Sections \ref{sec:consistency}-\ref{sec:corruptions}), the LF-CBM baseline is thresholded to match the average sparsity of HypCBM. This effectively binarizes the otherwise continuous activations into discrete active sets, ensuring a fair comparison.
All experiments are conducted on a single NVIDIA A100 GPU with curvature $c=0.1$, consistent with the pre-training configuration of the hyperbolic backbone, HyCoCLIP \cite{pal2024compositional}. 

\subsection{Classification Accuracy at Fixed Concept Budgets (ANEC)}
Standard accuracy obscures interpretability costs, as a model can score well using dense activations across all concepts. We therefore report \textit{Accuracy at Number of Effective Concepts} (ANEC) \cite{srivastava2025vlgcbmtrainingconceptbottleneck}, which measures performance under a strict budget of $K$ active concepts.
%Standard accuracy metrics often obscure the interpretability cost of CBMs, as models can maximize performance by utilizing dense concept activation vectors. To measure informational efficiency, we report \textit{Accuracy at Number of Effective Concepts} (ANEC) \cite{srivastava2025vlgcbmtrainingconceptbottleneck}, which measures performance under a strict budget of $K$ active concepts. 
We analyze the results in Table~\ref{tab:main_results} through three findings, and evaluate data efficiency across reduced training set sizes.

\paragraph{1. Better accuracy.} HypCBM consistently achieves higher peak performance than its Euclidean counterparts. Even when the bottleneck is relaxed ($K \to \text{all}$), HypCBM outperforms the standard LF-CBM baseline across all three datasets. For instance, on CIFAR-100, HypCBM reaches a peak accuracy of 72.2\%, surpassing LF-CBM (HyCoCLIP-20M, 69.5\%) by 2.7 percentage points and LF-CBM (CLIP-20M, 67.9\%) by over 4 percentage points. Since the backbone is identical in the former comparison, these gains are attributable directly to the geometric activation mechanism rather than representational capacity.

\paragraph{2. Sparsity leads to further gains.} While HypCBM performs best across the board, its advantages are most pronounced in the high-sparsity regime, critical for interpretability. When restricted to small concept budgets, the performance gap widens significantly; for instance, on ImageNet ($K=10$), HypCBM more than doubles the baseline accuracy (34.6\% vs. 15.3\%). These results rely on correct geometric alignment: applying a mismatched Euclidean activation mechanism to the hyperbolic backbone often degrades performance below the Euclidean baseline, showing the gains stem from our method as opposed to the backbone.
%Entailment-based activations therefore concentrate discriminative information far more efficiently than standard cosine similarity.

\paragraph{3. HypCBM compensates for scale.} HypCBM frequently matches or outperforms the OpenAI CLIP baseline, despite being trained on $20\times$ less data. For instance, on the hierarchical SUN397 benchmark at ANEC-20, HypCBM achieves 65.7\% accuracy, beating the Euclidean CLIP-400M baseline by nearly 7 percentage points (65.7\% vs. 58.8\%). HypCBM attains this performance with just 20 concepts, matching an accuracy level that the CLIP-400M baseline requires $\approx 70$ active concepts to reach. This demonstrates that entailment-based concept activations can compensate for substantial pre-training data disadvantages in sparse interpretable regimes.

\begin{wraptable}{r}{0.29\linewidth}
\vspace{-1.5em}
\centering
\caption{\textbf{Data Efficiency:} ANEC-20 at 1\% and 10\% training data, CIFAR-100.}
\vspace{-0.3em}
\label{tab:data_efficiency}
\small
\begin{tabular}{lcc}
\toprule
\textbf{Method} & \textbf{1\%} & \textbf{10\%} \\
\midrule
LF-CBM & 30.2 & 45.9 \\
\rowcolor{Gray}
\textbf{HypCBM} & \textbf{43.4} & \textbf{54.7} \\
\bottomrule
\end{tabular}
\vspace{-1em}
\end{wraptable}
\paragraph{4. Data efficiency.} HypCBM consistently outperforms LF-CBM 
across all training data regimes (Table~\ref{tab:data_efficiency}). With 
only 1\% of CIFAR-100 available as training data, HypCBM achieves an ANEC-20 of 43.4\%, surpassing LF-CBM by over 13 percentage points, with low variance across seeds confirming result stability. This demonstrates that entailment-based concept 
activations remain effective even under severe data constraints. Full 
accuracy-sparsity curves with standard deviations are provided in Appendix~\ref{app:data-efficiency}.

\subsection{Hierarchical Consistency}
\label{sec:consistency}
Beyond accuracy, interpretable models should exhibit logical coherence: if a fine-grained concept is active (\textit{`school'}), its parent (\textit{`building'}) must be too. We quantify \textbf{Hierarchical Consistency} as:
\begin{equation}
    \text{Consistency} = 1 - \frac{\sum \mathbf{1}[a_{\text{child}}=1 
    \land a_{\text{parent}}=0]}{\sum \mathbf{1}[a_{\text{child}}=1]}
\end{equation}
We evaluate on SUN397 and ImageNet using parent-child relations validated via WordNet, and use sparsity-matched baselines to prevent trivially achieving 100\% consistency by activating all concepts.

\paragraph{Results.} Table~\ref{tab:consistency} reveals a consistent pattern: HypCBM's relative advantage over the Euclidean baseline grows as $\eta_{\text{img}}$ tightens. At $\eta_{\text{img}}=1.2$ the baseline degrades sharply (0.08 on SUN397) while HypCBM more than doubles its consistency, confirming that entailment-based activations prioritize hierarchically valid concepts even under severe budget constraints.

To disentangle the two components of our approach, Appendix~\ref{app:consistency-extended} includes two ablations. First, replacing cosine similarity with our entailment activation on the \emph{identical} HyCoCLIP-20M backbone nearly doubles consistency at every threshold, isolating the activation mechanism as the primary driver of gains independent of backbone choice. Second, comparing against a hierarchy-aware Euclidean model \citep{alper2024hierarcaps} trained on 20$\times$ more data shows that HypCBM matches or exceeds it at the strictest budgets, suggesting the hyperbolic manifold contributes an inductive bias beyond what entailment-aligned training alone provides.

\begin{table}[t]
    \centering
    \caption{\textbf{Hierarchical Consistency.} HypCBM (HyCoCLIP-20M) consistently outperforms the Euclidean LF-CBM (CLIP-20M) baseline across validated WordNet pairs, with relative gains exceeding $+100\%$ at the strictest budget.}
    \label{tab:consistency}
    \resizebox{0.7\linewidth}{!}{
    \begin{tabular}{
        l
        r@{\hspace{6pt}}c@{\hspace{6pt}}c
        @{\hspace{14pt}}
        r@{\hspace{6pt}}c@{\hspace{6pt}}c
    }
        \toprule 
        & \multicolumn{3}{c}{\textbf{SUN397} ($N=1503$)} 
        & \multicolumn{3}{c}{\textbf{ImageNet} ($N=829$)} \\
        \cmidrule(lr){2-4} \cmidrule(lr){5-7} 
        $\eta_\text{img}$ 
        & LF-CBM & HypCBM & \textbf{Gain}
        & LF-CBM & HypCBM & \textbf{Gain} \\
        \midrule
        1.5 & 0.63 & \textbf{0.84} & \textbf{+32.5\%} 
            & 0.59 & \textbf{0.75} & \textbf{+27.9\%} \\
        1.4 & 0.43 & \textbf{0.54} & \textbf{+27.2\%} 
            & 0.42 & \textbf{0.63} & \textbf{+51.1\%} \\
        1.3 & 0.21 & \textbf{0.33} & \textbf{+58.9\%} 
            & 0.28 & \textbf{0.51} & \textbf{+80.4\%} \\
        1.2 & 0.08 & \textbf{0.17} & \textbf{+104.7\%} 
            & 0.16 & \textbf{0.33} & \textbf{+102.9\%} \\
        \bottomrule
    \end{tabular}
    }
    \vspace{-1.5em}
\end{table}
\begin{figure*}[b]
    \centering
    \vspace{-1em}
    \begin{subfigure}[b]{0.62\linewidth}
        \centering
        \includegraphics[width=\linewidth]{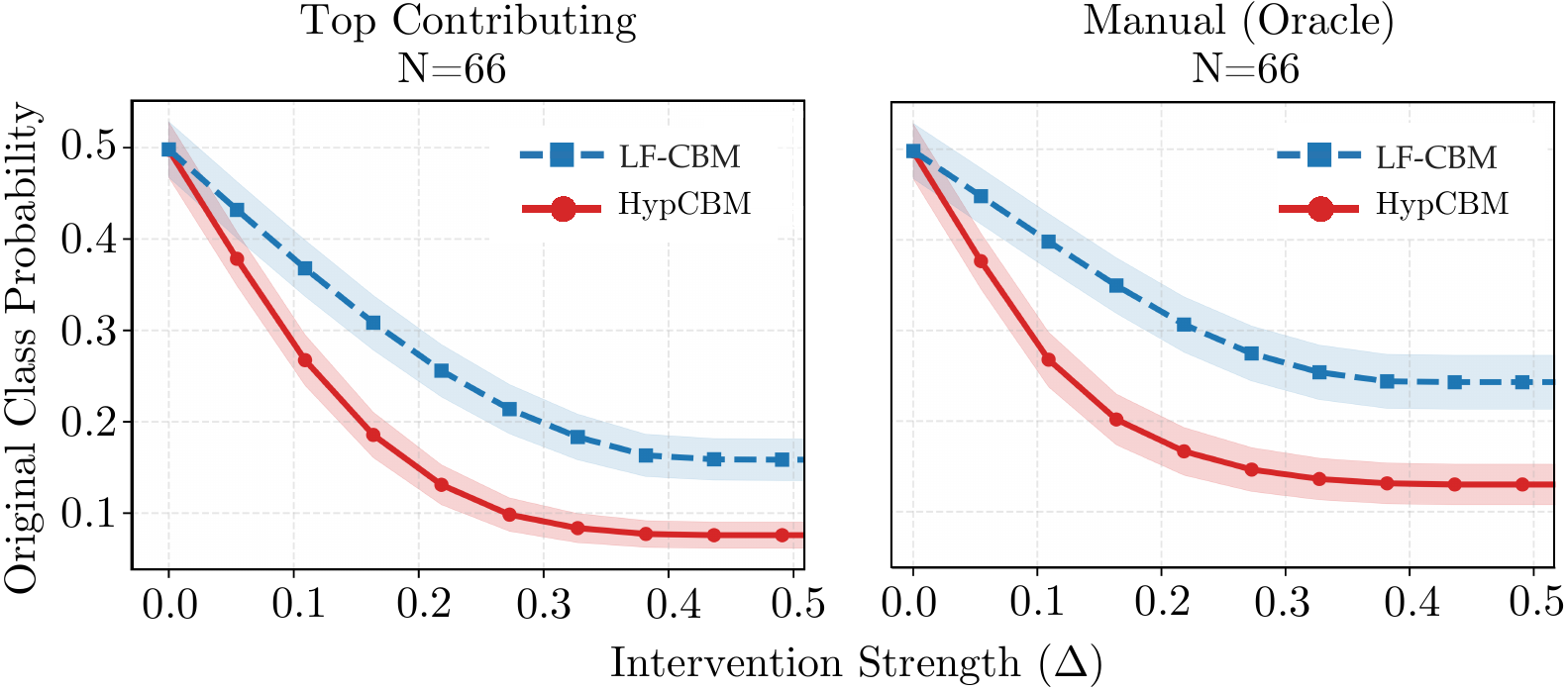}
        \caption{\textbf{Intervention Responsiveness.}}
        \label{fig:intervention}
    \end{subfigure}
    \hfill
    \begin{subfigure}[b]{0.36\linewidth}
        \centering
        \includegraphics[width=\linewidth]{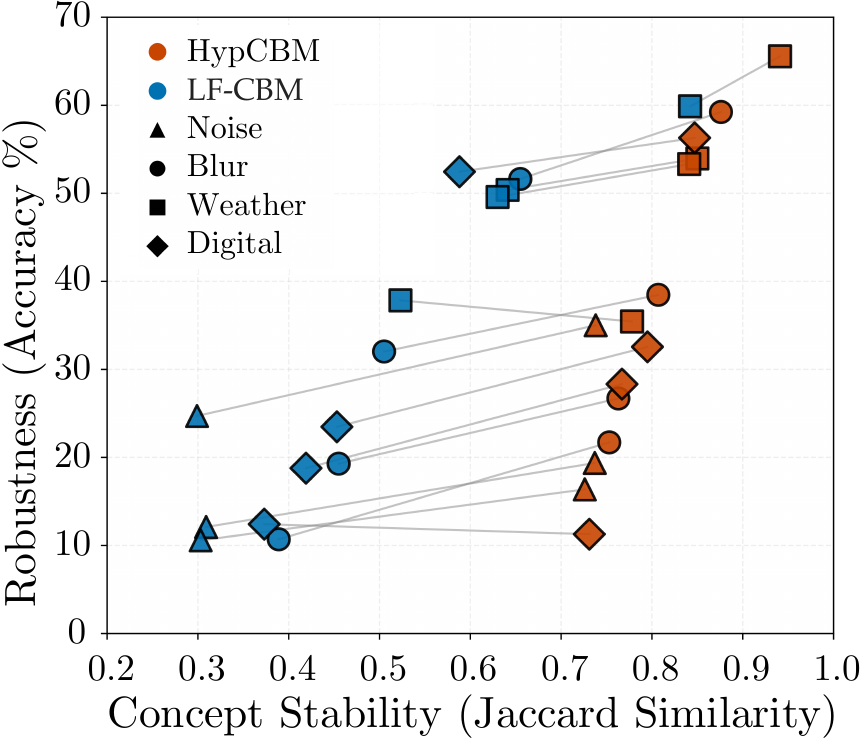}
        \caption{\textbf{Robustness and Concept Stability.}}
        \label{fig:stability}
    \end{subfigure}
    \vspace{-0.2em}
    \caption{\textbf{Intervention and Stability Analysis.} (a) HypCBM (red) displays steeper confidence decay than the Euclidean LF-CBM, indicating superior responsiveness to corrective interventions. Shaded bands show Standard Error (SE). (b) On CIFAR100-C, HypCBM maintains significantly higher semantic stability and accuracy across corruptions.}
\end{figure*}
\subsection{Hierarchically Faithful Concept Interventions}
\label{sec:interventionresults}
An essential part of any CBM is the ability to support human intervention. To quantify the utility of our hierarchical interventions, we simulate a human-in-the-loop correction scenario on misclassified test samples. For each incorrect prediction, we identify the active concepts contributing to the error and progressively suppress selected concepts by subtracting a scalar factor $\Delta$ from their activations. At each step, we measure the confidence assigned to the originally predicted (incorrect) class, yielding an intervention response curve.

We consider three strategies: \textbf{top-contributing} (suppressing the highest 
$w_i \cdot a_i$ concept), \textbf{manual oracle} (suppressing a top-contributing, ground-truth absent concept), and \textbf{random} (control baseline). We report results for top-contributing and manual strategies; random suppression serves as a control baseline. HypCBM and LF-CBM are evaluated at matched sparsity for fairness.

Fig.~\ref{fig:intervention} shows that HypCBM exhibits a steeper confidence decay than the Euclidean baseline, indicating stronger responsiveness to corrective edits. Under manual intervention, HypCBM successfully flips incorrect predictions to the correct class for 19\% more samples than LF-CBM, confirming the practical utility of hierarchically propagated interventions. Random interventions cause negligible confidence changes (Appendix~\ref{app:intervention}), confirming that observed gains stem from targeted suppression rather than general activation reduction. Crucially, augmenting LF-CBM with WordNet-derived propagation yields no measurable benefit: only $101/613$ CIFAR-100 concepts and $618/1935$ SUN397 concepts appear in WordNet, and none of the top contributing concepts have valid WordNet children. HypCBM's geometric propagation discovers implicit hierarchical relationships that taxonomy-based approaches cannot recover. 

\subsection{Robustness to Corruptions and Concept Stability} 
\label{sec:corruptions}
Robustness is a prerequisite for interpretability: explanations that change inconsistently under benign perturbations are unreliable, even if predictions remain correct. We therefore evaluate robustness at both the prediction and concept levels using CIFAR100-C \cite{hendrycks2019benchmarkingneuralnetworkrobustness}, which applies 15 types of corruption (e.g., noise, blur, weather) at varying severities.

Besides measuring \textbf{Prediction Robustness} (accuracy on corrupted data), we introduce \textbf{Concept Stability}: the semantic consistency of the model's explanations. To the best of our knowledge, no prior work has standardized a metric for explanation stability in CBMs, despite its importance for reliable human-in-the-loop interaction. We compute the Jaccard Similarity between the set of active concepts in the clean image ($S_{clean}$) and the corrupted image ($S_{corr}$):
\begin{equation}
    J(S_{clean}, S_{corr}) = \frac{|S_{clean} \cap S_{corr}|}{|S_{clean} \cup S_{corr}|}.
\end{equation}
HypCBM activations are naturally sparse and non-negative, allowing us to define concept activity via non-zero activation. We define the active concept sets $S$ using the sparsity-matched binarization protocol described at the start of this section. This ensures that differences in stability arise strictly from geometric structure rather than differing activation densities.

\paragraph{Results.} Fig.~\ref{fig:stability} highlights the resilience of our hyperbolic approach; the Euclidean LF-CBM baseline fails to maintain semantic consistency, with Jaccard similarity scores dropping as low as $0.3$ on noise corruptions. In contrast, HypCBM exhibits significantly higher concept stability, achieving an average Jaccard improvement of $+0.3$ over the baseline across corruption categories (see Appendix~\ref{app:robustness} for results at higher severities). Moreover, HypCBM achieves higher classification accuracy in 13 out of 15 corruption settings, confirming a strong association between stable semantic representations and robust predictive performance.

\section{Conclusion}
% In this work, we addressed a fundamental mismatch in the design of existing Concept Bottleneck Models: the representation of inherently hierarchical semantic concepts in flat Euclidean space. We argue that treating concepts as independent, orthogonal dimensions fundamentally misrepresents the taxonomic structure of human reasoning. By grounding concepts in hyperbolic geometry and replacing symmetric similarity with asymmetric geometric entailment, we demonstrate with HypCBM that proper geometric alignment leads to superior accuracy-sparsity trade-offs, data efficiency, robustness and stability. Crucially, this formulation enforces structural consistency by design, eliminating the logical paradoxes of prior methods and ensuring that model interventions respect the underlying semantic hierarchy. 
Existing Concept Bottleneck Models embed semantic concepts in flat Euclidean 
spaces, treating them as independent and orthogonal. We address this by 
grounding the concept bottleneck in hyperbolic geometry and replacing symmetric 
similarity with asymmetric geometric entailment. The resulting model, HypCBM, achieves concept activations that are hierarchically consistent by construction, without additional supervision or learned modules. The empirical results confirm that this geometric alignment translates into consistent gains in accuracy, data efficiency, robustness, and intervention quality, often rivaling models trained on substantially more data. Together, these findings suggest that structural priors on the representation space can serve as a powerful substitute for scale in sparse interpretable regimes.

Our findings highlight that interpretability is not merely a property of model outputs, but is shaped by the geometry of the representation space. Embedding concepts in a manifold that reflects their hierarchical relationships leads to explanations that are more accurate, robust to corruptions, and hierarchically consistent.
%HypCBM is a step towards geometrically grounded interpretability, where structural priors play a central role in bridging high-capacity foundation models and human-understandable reasoning.

\paragraph{Limitations and future work.} Our ablations partially isolate the contributions of representational geometry and activation design, but fully disentangling them remains an open question — for instance, through backbones that vary geometry while holding the training objective fixed. Additionally, HypCBM relies on taxonomic hierarchies derived from text (e.g., WordNet \cite{10.1145/219717.219748}), which do not always align with continuous visual semantics. In fine-grained settings such as CUB-200 \citep{WahCUB_200_2011}, discriminative cues are localized attributes (e.g., \textit{``spotted wing''}) rather than semantic parent-child relationships; incorporating visually grounded hierarchies is a promising direction. Finally, while HypCBM is backbone-agnostic, its performance depends on the quality of available hyperbolic VLMs. However, we also view this as an opportunity: as hyperbolic foundation models continue to advance \cite{huynh2026argentadaptivehierarchicalimagetext}, HypCBM is designed to scale with them naturally, providing a framework for translating structured representations into interpretable and controllable classifiers.

\bibliography{ref}
\bibliographystyle{plainnat}

\appendix
\onecolumn
\section{Additional Results}
\label{additional results}

\subsection{CUB200}
In this section, we provide a detailed analysis of HypCBM's performance on CUB-200-2011 \cite{WahCUB_200_2011}. Unlike the other datasets in this paper, which use LLM-generated concept banks, here we use the 312 expert-annotated visual attributes provided with the CUB-200-2011 dataset \cite{WahCUB_200_2011}, demonstrating that HypCBM generalizes beyond LLM-generated concept banks to pre-defined ones.

% \begin{figure}[H]
%         \centering
%         \includegraphics[width=0.5\linewidth]{figures/accuracy_vs_active_concepts_cub.pdf}
%         \label{fig:cub}
%     \caption{\textbf{Accuracy-Sparsity Trade-off, CUB-200.}}
%     \label{fig:tradeoff_combined}
% \end{figure}

\begin{table}[h]
    \centering
    \caption{\textbf{Additional Results.} Test accuracy at strict concept budgets ($K$) on CUB-200 using pre-defined expert-annotated attributes. We report peak accuracy and compare against the larger CLIP-400M backbone.}
    \resizebox{\linewidth}{!}{
    \begin{tabular}{lllccccccc}
    \toprule
        & & & \multicolumn{6}{c}{\textbf{Accuracy @ Concept Budget (ANEC-$K$)}} \\
        \cmidrule(lr){4-9}
        \textbf{Dataset} & \textbf{Method} & \textbf{Backbone} & \textbf{5} & \textbf{10} & \textbf{20} & \textbf{50} & \textbf{100} & \textbf{Peak} & \textbf{LP}  \\
        \midrule
        % \multirow{4}{*}{\shortstack[c]{\textbf{CIFAR-10}\\$(N_c=102)$}} 
        %  & PCBM & CLIP-20M & 73.87 & 81.66 & 83.90 & 88.21 & 88.71 & 89.46 \\
        %  & PCBM & HyCoCLIP-20M  & 81.17 & \textbf{88.60} & \textbf{89.60} & 90.97 & \textbf{92.95} & \textbf{93.08} \\
        %  & \cellcolor{Gray}\textbf{HypCBM (Ours)} & \cellcolor{Gray}HyCoCLIP-20M  & \cellcolor{Gray}\textbf{85.01} & \cellcolor{Gray}86.68 & \cellcolor{Gray}89.45 & \cellcolor{Gray}\textbf{91.25} & \cellcolor{Gray}90.88 & \cellcolor{Gray}91.00 \\
        %  \cmidrule{2-9}
        %  & \textit{Ref: PCBM} & \textit{CLIP-400M}  & \textit{\underline{88.95}} & \textit{\underline{90.65}} & \textit{\underline{92.00}} & \textit{\underline{92.97}} & \textit{\underline{93.93}} & \textit{\underline{94.24}}  \\
        % \midrule
        \multirow{4}{*}{\shortstack[c]{\textbf{CUB-200}\\$(N_c=317)$}} 
         & LF-CBM & CLIP-20M  & 6.4 & 11.0 & 17.7 & 30.8 & 40.7 & 41.8  & 56.5\\
         & LF-CBM & HyCoCLIP-20M  & 2.9 & 4.3 & 7.3 & 15.1 & 21.9 & 28.1 & 58.7\\
         & \cellcolor{Gray}\textbf{HypCBM (Ours)} & \cellcolor{Gray}HyCoCLIP-20M  & \cellcolor{Gray}\textbf{13.5} & \cellcolor{Gray}\textbf{20.8} & \cellcolor{Gray}\textbf{25.8} & \cellcolor{Gray}\textbf{36.8} & \cellcolor{Gray}\textbf{43.6} & \cellcolor{Gray}\textbf{45.3} & \cellcolor{Gray}58.7 \\
         \cmidrule{2-10}
         & \textit{Ref: LF-CBM} & \textit{CLIP-400M} & \textit{6.2} & \textit{16.4} & \textit{\underline{31.2}} & \textit{\underline{45.1}} & \textit{\underline{50.2}} & \textit{\underline{55.0}} & \textit{81.8} \\
        \bottomrule
    \end{tabular}
    }
    \label{tab:additionalresults}
\end{table}

\paragraph{CUB-200.} The CUB-200 dataset presents a challenging fine-grained classification task that relies heavily on localized features (e.g., 'beak shape', 'wing pattern'). In general, our multimodal approach is not well-suited for these kinds of scenarios, as two concepts like 'white undertail covert' and 'white undertail feather' might be represented very semantically similar, while they could be important discriminators. As shown in Table~\ref{tab:additionalresults}, the pre-trained HyCoCLIP and CLIP backbones (trained on 20M image-text pairs) struggle to capture these fine-grained distinctions compared to the OpenAI CLIP backbone (trained on 400M pairs). The standard Euclidean LF-CBM built on HyCoCLIP collapses under this representational gap, achieving a poor 2.86\% accuracy with 5 active concepts.

However, despite this poor performance, HypCBM demonstrates a remarkable ability to compensate for its weaker backbone. Most notably, at extremely low sparsity budgets (K=5 and K=10), HypCBM (trained on 20M data) surprisingly outperforms the LF-CBM CLIP-400M baseline (13.47\% vs. 6.23\% at K=5), suggesting that hyperbolic geometry allows for more efficient disentanglement of available semantic information in low-data regimes. As the concept budget relaxes ($K\geq20$), the raw feature quality of the 400M backbone eventually dominates, leading to a higher peak accuracy for the OpenAI baseline.

Another interesting result we observed, is the fact that accuracy does not keep increasing when more concepts are made available, but instead peaks at $\approx$ 150 active concepts and then degrades. This result may seem like a flaw at first, but is logically faithful; when classifying bird species, a model should not be able to use almost all of its available concepts. When it starts using false positive concepts this adds noise, hurting classification accuracy. In contrast, with the Euclidean baselines this happens only at the very end of the accuracy/sparsity curve, indicating information leakage; standard CBMs allow the model to assign small, non-zero weights to semantically irrelevant concepts, trading interpretability for the raw predictive power of the underlying dense embedding.

% \paragraph{CIFAR-10.} On CIFAR-10, HypCBM exhibits superior sparsity efficiency. It achieves 85.01\% accuracy with just 5 concepts, significantly outperforming the Euclidean PCBM on the same backbone (81.17\%) and the standard CLIP-20M baseline (73.87\%). While the Euclidean HyCoCLIP PCBM eventually catches up in peak accuracy, HypCBM remains the most efficient choice for highly constrained interpretability budgets, reaching near-peak performance faster than all 20M baselines.
\subsection{Hyperbolic Entailment Activation, CLIP, MERU \& HyCoCLIP}
\begin{table}[h]
    \centering
    \caption{\textbf{Ablation of Backbone Geometry.} We evaluate HypCBM on SUN397 using different pre-trained backbones. 
    CLIP embeddings are mapped to $\mathbb{L}^n$ via the exponential map.
    HyCoCLIP substantially outperforms MERU and CLIP, particularly in the sparse regime ($K \leq 20$), confirming the necessity of entailment-aligned embeddings.}
    \label{tab:hypcbm_meru}
    \begin{tabular}{lllcccccc}
    \toprule
       & & & \multicolumn{6}{c}{\textbf{Accuracy @ Concept Budget (ANEC-$K$)}} \\
       \cmidrule(lr){4-9}
       \textbf{Dataset} & \textbf{Method} & \textbf{Backbone} & \textbf{5} & \textbf{10} & \textbf{20} & \textbf{50} & \textbf{100} & \textbf{Peak}  \\
       \midrule
       \multirow{3}{*}{\shortstack[c]{\textbf{SUN397}\\$(N_c=1935)$}} 
        & HypCBM & CLIP-20M   & 32.3 & 35.9 & 42.9 & 51.7 & 54.2 & 67.3 \\
         & HypCBM & MERU-20M   & 23.1 & 28.3 & 37.5 & 51.1 & 58.5 & 69.7 \\
         & \cellcolor{Gray}HypCBM & \cellcolor{Gray}HyCoCLIP-20M   & \cellcolor{Gray}\textbf{49.3} & \cellcolor{Gray}\textbf{58.3} & \cellcolor{Gray}\textbf{65.7} & \cellcolor{Gray}\textbf{69.9} & \cellcolor{Gray}\textbf{71.3} & \cellcolor{Gray}\textbf{72.3} \\
       \bottomrule
    \end{tabular}
\end{table}

\paragraph{Results.} 
Table~\ref{tab:hypcbm_meru} isolates the impact of embedding geometry. We observe that mapping standard CLIP features to $\mathbb{L}^n_c$ fails to create a functional hierarchy: because CLIP features are L2-normalized, they map to a fixed distance from the origin, resulting in constant (very small) cone apertures for all concepts. Without radial variance to distinguish generic from specific attributes, the entailment condition collapses, forcing the model to rely solely on angular separation. Furthermore, while MERU is natively hyperbolic, it significantly underperforms in the sparse regime ($K \leq 20$), suggesting that hyperbolic geometry alone is insufficient. While MERU enforces image-text entailment, it does not explicitly train concept-to-concept partial orders; HyCoCLIP's compositional cone training ensures that concept embeddings are radially ordered by specificity, which is the structure HypCBM's activation mechanism depends on.

The results in Table~\ref{tab:hypcbm_meru} are achieved by significantly inflating the strictness parameter $\eta_{img}$ to about twice the value used for HyCoCLIP, forcing most concepts to be activated by effectively reverting the model to a standard Euclidean LF-CBM (with non-negative activations) based mostly on angular separation, negating the benefits of the hierarchy.

\subsection{Sparsity-Matched Evaluation}
\label{app:sparsity-matched}

A potential confound in ANEC comparisons is that HypCBM's entailment-based 
activations are naturally sparse, whereas LF-CBM produces dense continuous 
activations. To verify that HypCBM's gains stem from geometrically meaningful 
activations rather than sparsity alone, we evaluate a sparsity-matched variant 
of LF-CBM (LF-CBM Matched) in which only the top-$K$ highest concept 
activations are retained, matching HypCBM's average active concept count at 
each budget.
\begin{table}[h]
    \centering
    \caption{\textbf{Sparsity-Matched Evaluation.} ANEC-$K$ accuracy when 
    LF-CBM activations are thresholded to retain only the top-$K$ highest 
    activations (LF-CBM Matched), controlling for sparsity. Enforcing 
    sparsity on LF-CBM consistently degrades performance, confirming that 
    HypCBM's gains arise from geometrically meaningful activations rather 
    than sparsity alone.}
    \label{tab:sparsity_matched}
    \resizebox{0.8\linewidth}{!}{
    \begin{tabular}{llccccc}
        \toprule
        & & \multicolumn{5}{c}{\textbf{Sparsity Matched Accuracy @ K}} \\
        \cmidrule(lr){3-7}
        \textbf{Dataset} & \textbf{Method} & \textbf{5} & \textbf{10} 
        & \textbf{20} & \textbf{50} & \textbf{100} \\
        \midrule
        \multirow{4}{*}{\textbf{CIFAR-100}}
          & LF-CBM (CLIP-20M)          & 39.7 & 48.9 & 55.4 & 60.6 & 64.5 \\
          & LF-CBM Matched (CLIP-20M)  & 25.0 & 39.1 & 55.1 & 61.3 & 63.1 \\
          & \cellcolor{Gray}\textbf{HypCBM (HyCoCLIP-20M)} 
            & \cellcolor{Gray}\textbf{44.3} & \cellcolor{Gray}\textbf{54.4} 
            & \cellcolor{Gray}\textbf{60.9} & \cellcolor{Gray}\textbf{69.7} 
            & \cellcolor{Gray}\textbf{71.8} \\
          \cmidrule{2-7}
          & \textit{Ref: LF-CBM (CLIP-400M)} 
            & \textit{28.0} & \textit{58.0} & \textit{63.9} 
            & \textit{64.7} & \textit{67.0} \\
        \midrule
        \multirow{4}{*}{\textbf{SUN397}}
          & LF-CBM (CLIP-20M)          & 45.8 & 53.4 & 61.3 & 67.0 & 68.6 \\
          & LF-CBM Matched (CLIP-20M)  & 42.3 & 50.2 & 58.4 & 67.2 & 68.7 \\
          & \cellcolor{Gray}\textbf{HypCBM (HyCoCLIP-20M)} 
            & \cellcolor{Gray}\textbf{49.3} & \cellcolor{Gray}\textbf{58.3} 
            & \cellcolor{Gray}\textbf{65.7} & \cellcolor{Gray}\textbf{69.9} 
            & \cellcolor{Gray}\textbf{71.3} \\
          \cmidrule{2-7}
          & \textit{Ref: LF-CBM (CLIP-400M)} 
            & \textit{40.0} & \textit{48.0} & \textit{58.8} 
            & \textit{64.4} & \textit{66.8} \\
        \bottomrule
    \end{tabular}
    }
\end{table}

Table~\ref{tab:sparsity_matched} shows that enforcing sparsity on LF-CBM 
consistently \emph{hurts} performance, particularly at low budgets (e.g., 
CIFAR-100 at $K=5$: 39.7\% $\to$ 25.0\%). This confirms that the gains of 
HypCBM are not an artifact of activation sparsity: if sparsity were the 
primary driver, enforcing it on LF-CBM would help rather than hurt. Instead, 
HypCBM's sparse activations are geometrically meaningful — the concepts that 
activate are those for which the image genuinely falls within the entailment 
cone, whereas forcing sparsity on cosine similarity activations discards 
informative but lower-ranked concepts. HypCBM outperforms LF-CBM Matched 
by a wide margin across all budgets and both datasets.
\newpage
\subsection{Additional Intervention Results}
\label{app:intervention}
\begin{figure}[h]
    \centering
    \includegraphics[width=\linewidth]{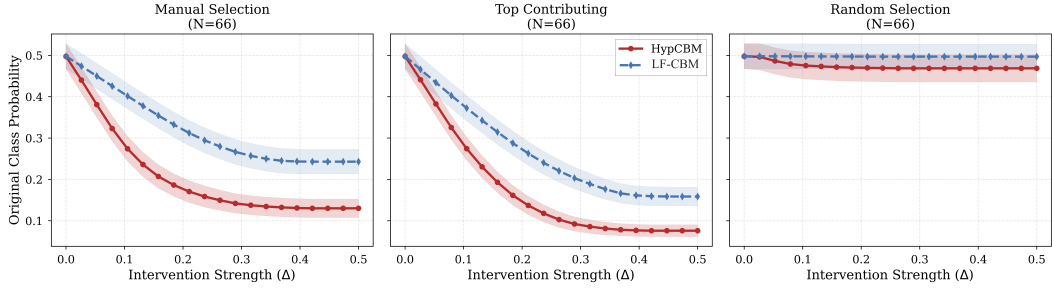}
    \caption{\textbf{Complete Intervention Analysis.} We compare the probability response of HypCBM (Red) and LF-CBM (Blue) across three distinct strategies. \textbf{Left (Manual):} We intervene on the concept with the highest contribution that is ground-truth absent (a false positive). HypCBM displays the sharpest confidence decay, indicating it is highly responsive to valid human corrections. \textbf{Center (Top-Contributing):} We intervene on the single highest contributing concept. The trend mirrors the manual setting, confirming utility for automated debugging. \textbf{Right (Random):} We intervene on random active concepts. Both models show negligible response, serving as a negative control; this confirms that the performance drops observed in the other settings are driven by semantic relevance, not model instability.} \label{fig:app_interventions}
    \label{fig:placeholder}
\end{figure}

\subsection{Data Efficiency}
\label{app:data-efficiency}
\begin{figure}[h!]
    \centering
    \hspace*{-1.2em}\includegraphics[width=0.6\linewidth]{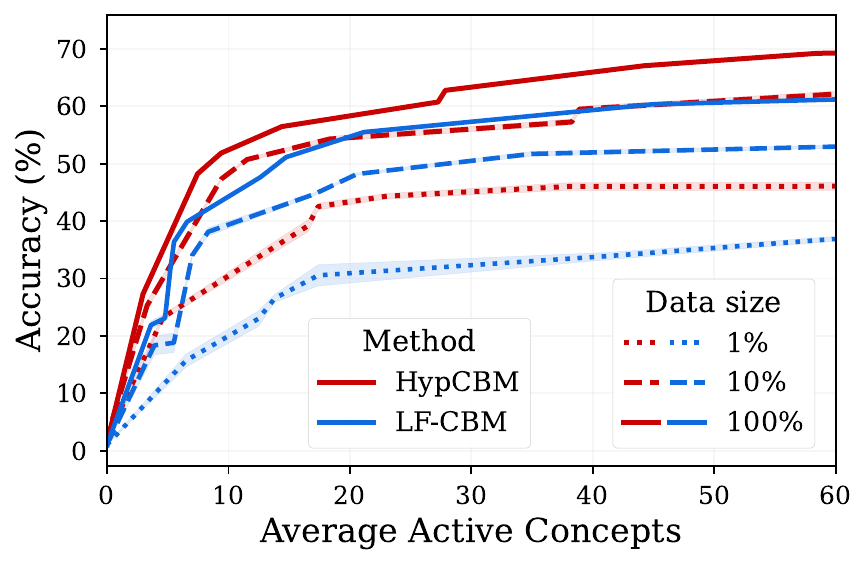}
    \caption{\textbf{Data efficiency on CIFAR100.} HypCBM outperforms LF-CBM (CLIP-20M) for any data budget.}
    \label{fig:data_efficiency}
    \vspace{-1em}
\end{figure}
Figure~\ref{fig:data_efficiency} shows ANEC curves on CIFAR-100 across training set sizes, averaged over 3 seeds with variance bands. A striking cross-regime comparison emerges: HypCBM trained on just 10\% of the data matches LF-CBM trained on the full dataset across all concept budgets, demonstrating a 10$\times$ data efficiency advantage. At 100\% training data, variance across seeds is negligible for both methods (std at ANEC-20: 0.09\% for HypCBM, 0.13\% for LF-CBM), confirming that main results are stable. At 1\% data, LF-CBM shows substantially higher variance whereas HypCBM remains comparatively stable, suggesting that entailment-based geometric activations provide a more reliable inductive bias precisely when supervision is scarce.

\subsection{Additional Hierarchical Consistency Results}
\label{app:consistency-extended}
Table~\ref{tab:combined_consistency} extends the hierarchical consistency 
analysis from the main paper by including two additional baselines: 
LF-CBM with a large-scale CLIP backbone (CLIP-400M) and a hierarchy-aware 
Euclidean CLIP model (CLIP$^{\text{RE}}$-400M \citep{alper2024hierarcaps}), 
both trained on $20\times$ more data than HyCoCLIP. From this experiment, we gain three insights.

\begin{table}[h]
    \centering
    \caption{\textbf{Hierarchical Consistency, SUN397 \& ImageNet.} We measure the rate at which activating a child concept implies activation of its parent across $N$ validated WordNet pairs. Despite a massive $20\times$ pre-training data disadvantage, HypCBM significantly outperforms both standard CLIP and the hierarchy-aware CLIP$^{\text{RE}}$ at the strict margins ($\eta \le 1.3$) required for interpretability.}
    \label{tab:combined_consistency}
    \resizebox{\linewidth}{!}{
    \begin{tabular}{@{\hspace{1pt}}l@{\hspace{12pt}} c@{\hspace{8pt}}c@{\hspace{8pt}}c@{\hspace{8pt}}c@{\hspace{12pt}} c@{\hspace{-1pt}} c@{\hspace{8pt}}c@{\hspace{8pt}}c@{\hspace{8pt}}c@{\hspace{0.5pt}}}
        \toprule
         & \multicolumn{4}{c}{\textbf{SUN397} ($N=1503$)} & & \multicolumn{4}{c}{\textbf{ImageNet} ($N=829$)} \\
        \cmidrule(lr){2-5} \cmidrule(lr){7-10}
         & LF-CBM & LF-CBM & LF-CBM & \textbf{HypCBM (Ours)} & & LF-CBM & LF-CBM & LF-CBM & \textbf{HypCBM (Ours)} \\
        $\eta_\text{{img}}$ & \scriptsize{(CLIP-20M)} & \scriptsize{(CLIP-400M)} & \scriptsize{(CLIP$^{RE}$-400M)} & \scriptsize{\textbf{(HyCoCLIP-20M)}} & & \scriptsize{(CLIP-20M)} & \scriptsize{(CLIP-400M)} & \scriptsize{(CLIP$^{RE}$-400M)} & \scriptsize{\textbf{(HyCoCLIP-20M)}} \\
        \midrule
        1.5  & 0.63 & 0.63 & 0.80 & \textbf{0.84} & & 0.59 & 0.70 & \textbf{0.82} & 0.75 \\
        1.4  & 0.43 & 0.46 & \textbf{0.62} & 0.54 & & 0.42 & 0.56 & \textbf{0.68} & 0.63 \\
        1.3  & 0.21 & 0.30 & \textbf{0.36} & 0.33 & & 0.28 & 0.41 & 0.48 & \textbf{0.51} \\
        1.2  & 0.08 & 0.14 & 0.14 & \textbf{0.17} & & 0.16 & 0.22  & 0.20 & \textbf{0.33} \\
        \bottomrule
    \end{tabular}
    }
\vspace{-0.5em}
\end{table}

\paragraph{Data scale helps for consistency.} Scaling from CLIP-20M to 
CLIP-400M yields consistent but modest gains across all thresholds 
(e.g., 0.63 $\to$ 0.63 at $\eta=1.5$ on SUN397, 0.08 $\to$ 0.14 at 
$\eta=1.2$). This confirms that data scale alone is insufficient to 
close the gap with HypCBM.

\paragraph{Hierarchical training helps more, but geometry still wins at 
strict budgets.} CLIP$^{\text{RE}}$, which adds an entailment-aware 
training objective on top of the larger dataset, improves substantially 
over standard CLIP-400M, particularly at loose thresholds ($\eta \geq 1.4$). 
However, the advantage narrows at stricter thresholds: at $\eta=1.2$, 
HypCBM matches or exceeds CLIP$^{\text{RE}}$ on both datasets (SUN397: 
0.17 vs.\ 0.14; ImageNet: 0.33 vs.\ 0.20), despite operating with $20\times$ 
less pre-training data. This crossover suggests that the hyperbolic manifold 
provides an inductive bias for hierarchical consistency that entailment-aligned 
training in Euclidean space cannot fully replicate.

\begin{table}[h]
    \centering
    \caption{\textbf{Hierarchical Consistency: Activation Mechanism Ablation.} Using the \emph{identical} HyCoCLIP-20M backbone on SUN397, replacing standard cosine similarity (LF-CBM) with our entailment activation (HypCBM) nearly doubles consistency, isolating the geometric activation mechanism as the primary source of the performance gains.}
    \label{tab:consistency_ablation}
    \begin{tabular}{lcc}
        \toprule
        & \multicolumn{2}{c}{\textbf{SUN397 (HyCoCLIP-20M)}} \\
        \cmidrule(lr){2-3}
        $\eta_\text{img}$ & LF-CBM & HypCBM \\
        \midrule
        1.5 & 0.44 & \textbf{0.84} \\
        1.4 & 0.26 & \textbf{0.54} \\
        1.3 & 0.12 & \textbf{0.33} \\
        1.2 & 0.05 & \textbf{0.17} \\
        \bottomrule
    \end{tabular}
    \vspace{-0.5em}
\end{table}

\paragraph{The activation mechanism is the primary driver.} 
Table~\ref{tab:consistency_ablation} isolates the contribution of our 
entailment activation by holding the backbone fixed. Using the identical 
HyCoCLIP-20M backbone, replacing cosine similarity with our margin-of-inclusion 
activation nearly doubles consistency at every threshold (e.g., 0.44 $\to$ 
0.84 at $\eta=1.5$, 0.05 $\to$ 0.17 at $\eta=1.2$ on SUN397). Since the 
backbone is identical, these gains are exclusively attributable to the activation mechanism. Combined with the CLIP$^{\text{RE}}$ comparison above, the evidence suggests that both components are necessary: hyperbolic geometry provides the representational structure, and the margin-of-inclusion activation reads it out effectively at inference time.

\begin{table*}[h]
    \centering
    \caption{\textbf{Main Results.} Test accuracy at strict concept budgets ($K$). HypCBM consistently outperforms Euclidean baselines (LF-CBM) on identical backbones (HyCoCLIP-20M) and often even baselines that use large-scale pre-trained backbones (CLIP-400M and CLIP$^{RE}$-400M).}
    \begin{tabular}{lllcccccc}
    \toprule
       & & & \multicolumn{6}{c}{\textbf{Accuracy @ Concept Budget (ANEC-$K$)}} \\
       \cmidrule(lr){4-9}
       \textbf{Dataset} & \textbf{Method} & \textbf{Backbone} & \textbf{5} & \textbf{10} & \textbf{20} & \textbf{50} & \textbf{100} & \textbf{Peak}  \\
       \midrule
       \multirow{5}{*}{\shortstack[c]{\textbf{CIFAR-100}\\$(N_c=613)$}} 
         & LF-CBM & CLIP-20M & 39.7 & 48.9 & 55.4 & 60.6 & 64.5 & 67.9 \\
         & LF-CBM & HyCoCLIP-20M  & 25.6 & 34.6 & 56.3 & 61.6 & 62.6 & 69.5 \\
         & \cellcolor{Gray}\textbf{HypCBM (Ours)} & \cellcolor{Gray}HyCoCLIP-20M  & \cellcolor{Gray}\textbf{44.3} & \cellcolor{Gray}\textbf{54.4} & \cellcolor{Gray}\textbf{60.9} & \cellcolor{Gray}\textbf{69.7} & \cellcolor{Gray}\textbf{71.8} & \cellcolor{Gray}\textbf{72.2} \\
         \cmidrule{2-9}
         & \textit{Ref: LF-CBM} & \textit{CLIP-400M}  & \textit{28.0} & \textit{\underline{58.0}} & \textit{\underline{63.9}} & \textit{64.7} & \textit{67.0} & \textit{71.3}  \\
         & \textit{Ref: LF-CBM} & \textit{CLIP$^{RE}$-400M} & \textit{20.4} & \textit{26.0} & \textit{31.7} & \textit{46.6} & \textit{58.3} & \textit{70.2} \\
       \midrule
       \multirow{5}{*}{\shortstack[c]{\textbf{SUN397}\\$(N_c=1935)$}} 
         & LF-CBM & CLIP-20M         & 45.8 & 53.4 & 61.3 & 67.0 & 68.6 & 70.5 \\
         & LF-CBM & HyCoCLIP-20M   & 34.3 & 44.2 & 50.4 & 59.1 & 61.5 & 72.1 \\
         & \cellcolor{Gray}\textbf{HypCBM (Ours)} & \cellcolor{Gray}HyCoCLIP-20M   & \cellcolor{Gray}\textbf{49.3} & \cellcolor{Gray}\textbf{58.3} & \cellcolor{Gray}\textbf{65.7} & \cellcolor{Gray}\textbf{69.9} & \cellcolor{Gray}\textbf{71.3} & \cellcolor{Gray}\textbf{72.3} \\
         \cmidrule{2-9}
         & \textit{Ref: LF-CBM} & \textit{CLIP-400M} & \textit{40.0} & \textit{48.0} & \textit{58.8} & \textit{64.4} & \textit{66.8} & \textit{\underline{79.0}} \\
         & \textit{Ref: LF-CBM} & \textit{CLIP$^{RE}$-400M} & \textit{26.9} & \textit{34.4} & \textit{40.1} & \textit{51.5} & \textit{59.9} & \textit{73.2} \\
       \bottomrule
    \end{tabular}
    \label{tab:clipreacc}
\end{table*}

For reference, Table~\ref{tab:clipreacc} reports accuracy for the hierarchy-aware Euclidean CLIP variant CLIP$^{\text{RE}}$ \cite{alper2024hierarcaps}\footnote{CLIP$^{\text{RE}}$ features are L2-normalized prior to computing cosine similarity.}. A notable pattern emerges: despite being trained on 20$\times$ more data, CLIP$^{RE}$ performs substantially worse than standard CLIP-20M at strict concept budgets (e.g., SUN397 at K=5: 26.9\% vs.\ 45.8\%), suggesting that Euclidean entailment training trades sparse accuracy for hierarchical structure. HypCBM avoids this trade-off by grounding structure in the geometry of the manifold rather than the training objective, achieving both stronger consistency (Table~\ref{tab:combined_consistency}) and higher accuracy at low budgets.

\subsection{Additional Robustness Results (CIFAR100-C)}
\label{app:robustness}

In the main text, we analyzed concept stability at Severity 1. To ensure these findings are consistent across varying levels of image degradation, we evaluated both models across the full range of corruption severities (1--5) in the CIFAR-100-C benchmark.

\begin{figure}[h]
    \centering
    % Replace 'stability_collapse_curve.pdf' with your actual filename
    \includegraphics[width=0.5\linewidth]{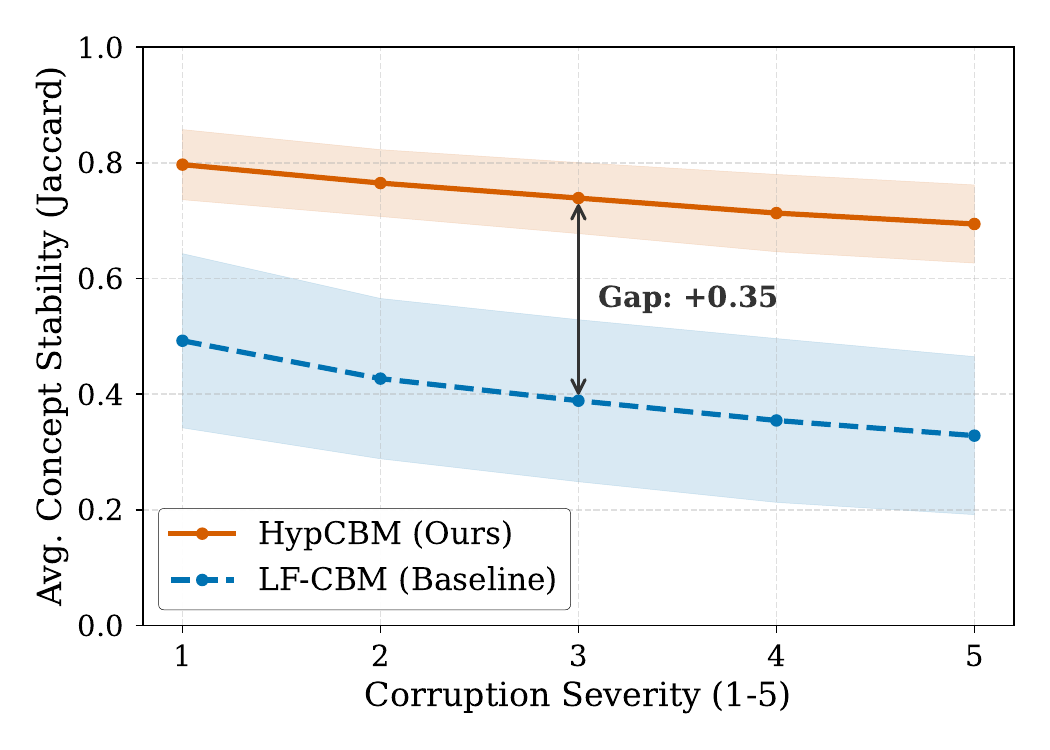}
    \caption{\textbf{Semantic stability across five severities of input corruption.} The plot shows the average Jaccard similarity across all 15 corruption types as a function of severity. The shaded regions represent one standard deviation. The baseline (LF-CBM) shows a rapid decrease in concept stability as severity increases, while HypCBM maintains high stability ($J > 0.7$) even at Severity 5. The stability gap remains consistently large ($\sim0.35$) throughout.}
    \label{fig:collapse_curve}
    \vspace{-1em}
\end{figure}

Figure \ref{fig:collapse_curve} presents the aggregated results. We observe a consistent gap between the hyperbolic and Euclidean models. Notably, at Severity 3 (the tipping point for accuracy), the gap in concept stability is approximately $0.35$, confirming that HypCBM's explanations are significantly more robust to input perturbations.

\subsection{Hyperbolic Norm Filtering}
\label{sec:ablation}
We evaluate the effect of filtering low-norm concepts, as introduced in Section~\ref{sec:method}, on the SUN397 dataset. In hyperbolic representations, embedding norm correlates with specificity; points near the origin represent broad, generic concepts (e.g., "object") with naturally wide entailment cones.

\begin{figure}[h]
    \centering
    \includegraphics[width=0.6\linewidth]{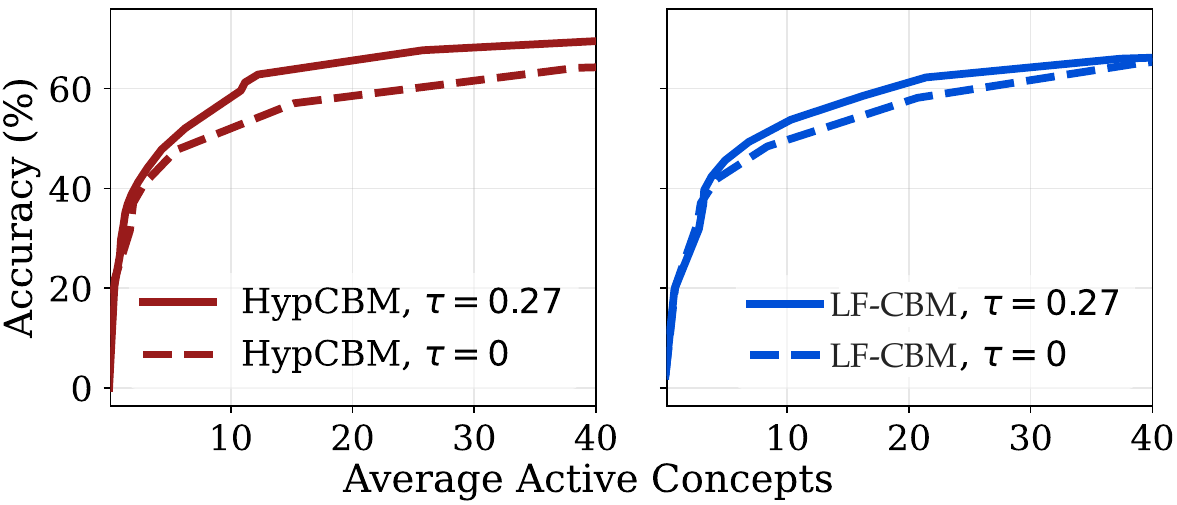}
    \caption{\textbf{Ablation on Hyperbolic Norm Filtering, SUN397. $\tau=0.27$.}}
    \label{fig:normfiltering}
\end{figure}

As shown in Figure~\ref{fig:normfiltering}, filtering these generic concepts ($\tau=0.27$) lifts the accuracy-sparsity frontier for both models, with HypCBM profiting disproportionally. Because our entailment mechanism links cone width to embedding norm, concepts near the origin have excessively wide apertures that activate indiscriminately, hurting efficiency. Filtering these trivial activations allows HypCBM to recover substantial gains. The Euclidean LF-CBM, in contrast, relies on angular similarity; pruning general concepts does not lead to a comparable reduction in trivial activations.

\section{Concept Bank}
\label{app:concept bank}
As the concept banks for CIFAR100, ImageNet and CUB-200 were already created and made public by \citet{oikarinen2023labelfree}, we only apply the concept bank creation process for SUN397. We use the following prompts for each class in the dataset, as introduced by \citet{oikarinen2023labelfree}:
\begin{figure*}[h]
    \centering
    \small % Adjust font size for the whole block if needed
    % --- Column 1: Important Features ---
    \begin{minipage}[t]{0.32\linewidth}
        \textbf{1. Important Features} \\
        
        \scriptsize % Smaller font for the code
        \texttt{"List the most important features for recognizing something as a 'goldfish':\\
        - bright orange color\\
        - a small, round body\\
        - a long, flowing tail\\
        - a small mouth\\
        - orange fins\\
        \\
        List the most important features for recognizing something as a 'beer glass':\\
        - a tall, cylindrical shape\\
        - clear or translucent color\\
        - opening at the top\\
        - a sturdy base\\
        - a handle\\
        \\
        List the most important features for recognizing something as a '\{\}':"}
    \end{minipage}
    \hfill % Adds spacing between columns
    % --- Column 2: Superclass ---
    \begin{minipage}[t]{0.32\linewidth}
        \textbf{2. Superclass} \\
        
        \scriptsize
        \texttt{"Give superclasses for the word 'tench':\\
        - fish\\
        - vertebrate\\
        - animal\\
        \\
        Give superclasses for the word 'beer glass':\\
        - glass\\
        - container\\
        - object\\
        \\
        Give superclasses for the word '\{\}':"}
    \end{minipage}
    \hfill
    % --- Column 3: Context (Around) ---
    \begin{minipage}[t]{0.32\linewidth}
        \textbf{3. Context (Around)} \\
        
        \scriptsize
        \texttt{"List the things most commonly seen around a 'tench':\\
        - a pond\\
        - fish\\
        - a net\\
        - a rod\\
        - a reel\\
        - a hook\\
        - bait\\
        \\
        List the things most commonly seen around a 'beer glass':\\
        - beer\\
        - a bar\\
        - a coaster\\
        - a napkin\\
        - a straw\\
        - a lime\\
        - a person\\
        \\
        List the things most commonly seen around a '\{\}':"}
    \end{minipage}
    
    \caption{\textbf{LLM Prompts for Concept Generation.} We use three distinct prompt templates (Important Features, Superclass, Context) to generate diverse visual attributes. These few-shot examples are fed to GPT-3 to produce the raw concept bank.}
    \label{fig:prompts}
\end{figure*}

After this initial set of candidate concepts is created, a few processing steps are applied. First, concepts that are too long (longer than 30 tokens) are removed. Then, we calculate cosine similarity between all class labels and all concepts and remove those with a similarity $>0.85$. Finally, the same thing is done for the concepts themselves; any concept with a similarity of $>0.9$ to another concept is removed (the concept with lower average similarity to other concepts is deleted). This process results in a different concept bank size, $N_c$, for each dataset, mostly dependent on how many classes they contain, and how much redundancy across generated concepts is present.

\section{Calibration of Entailment Strictness Parameters $\eta_{text}$ and $\eta_{img}$.}
\label{app:eta_img}

\subsection{Finding the Optimal Inter-Modal $\eta_{img}$}
Because standard image datasets lack exhaustive, multi-level hierarchical annotations, we leverage the fundamental premise of aligned multimodal spaces: semantic entailment should be modality-agnostic. In a well-aligned shared manifold, if a specific text (e.g., "Persian cat") entails a general text ("cat"), an image depicting that specific concept should geometrically entail the general concept as well. Thus, we utilize WordNet's text-text hypernym chains as a proxy to calibrate the image-text threshold $\eta_{img}$.

To determine a robust value for the entailment strictness parameter $\eta_{img}$, we treat the calibration as a binary classification problem: distinguishing "true" semantic entailments from non-entailing pairs based on their geometric entailment ratio:
\[R=\frac{\phi(\mathbf{z},\mathbf{c})}{\omega(\mathbf{c})}.\] 
Since $\eta_{img}$ acts as the decision boundary ($R \leq \eta_{img}$), we seek the threshold that optimally separates these two classes. We construct the following two proxy datasets:
\begin{itemize}
    \item \textbf{Positive Set}: A set of ground-truth entailment pairs derived from WordNet hypernym chains (e.g., Persian cat $\rightarrow$ cat), filtered to ensure the parent concept has a lower hyperbolic norm than the child.
    \item \textbf{Negative Set}: A set of randomly sampled concept pairs with no known semantic relationship, serving as a baseline for non-entailment.
\end{itemize}

Then, we compute the entailment ratio R for all pairs in both sets. We then sweep across possible values of $\eta_{img}$ to calculate the True Positive Rate (Sensitivity, or the fraction of WordNet pairs correctly identified as entailed ($R\leq \eta_{img}$)) and False Positive Rate (1-Specificity, the fraction of random pairs correctly identified as not entailed ($R>\eta_{img}$)) at each threshold.

To select the optimal $\eta_{img}$, we maximize Youden’s J statistic (J):
\[J(\eta_{img})=\text{Sensitivity}(\eta_{img}) + \text{Specificity}(\eta_{img}) - 1\]

\begin{figure}[htbp]
    \centering
    % First Subfigure
    \begin{subfigure}[b]{0.48\textwidth}
        \centering
        % Adjusted width to fit inside the subfigure
        \includegraphics[width=\linewidth]{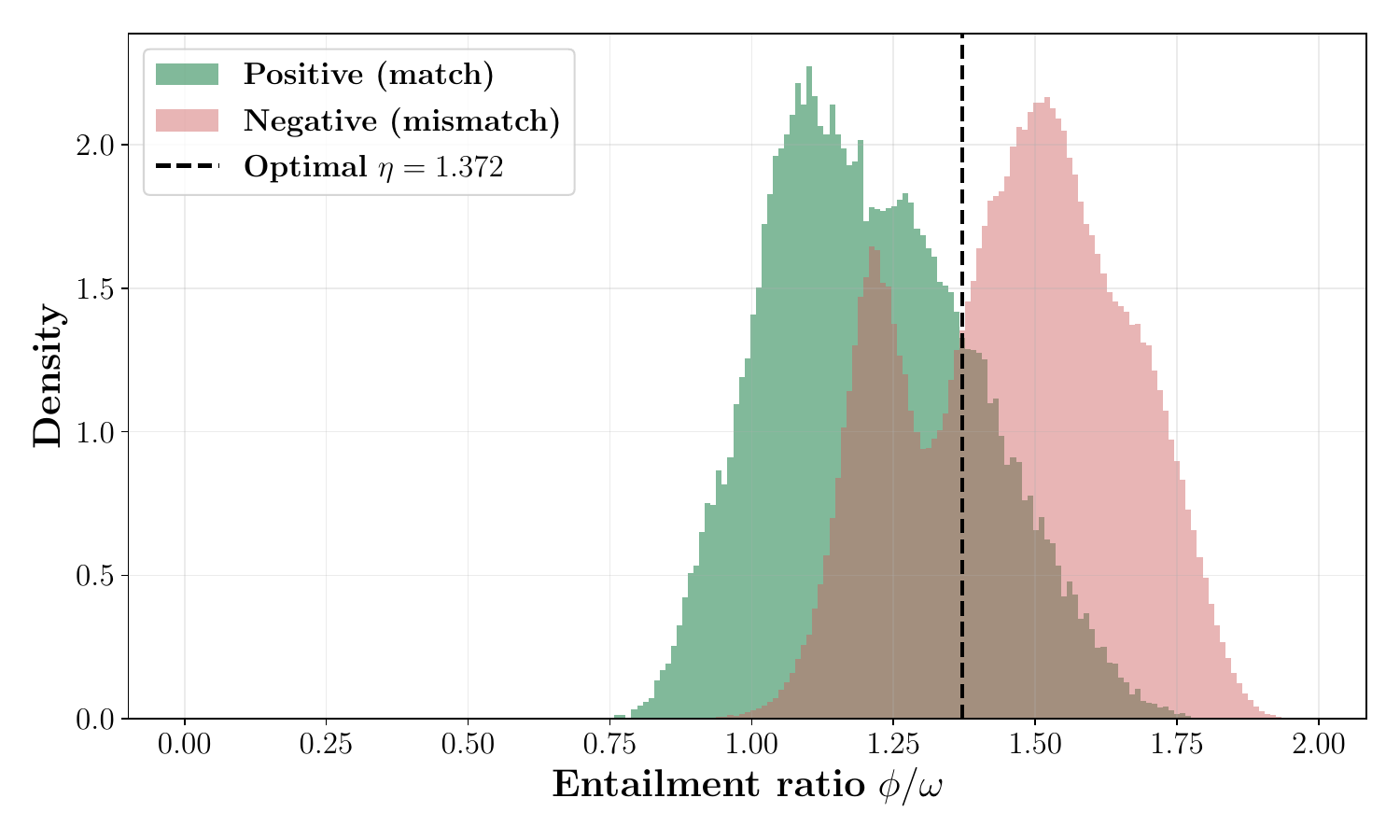}
        \caption{CIFAR100}
        \label{fig:cifar100}
    \end{subfigure}
    % \hfill % Adds flexible space between the images
    % Second Subfigure
    \begin{subfigure}[b]{0.48\textwidth}
        \centering
        % Adjusted width to fit inside the subfigure
        \includegraphics[width=\linewidth]{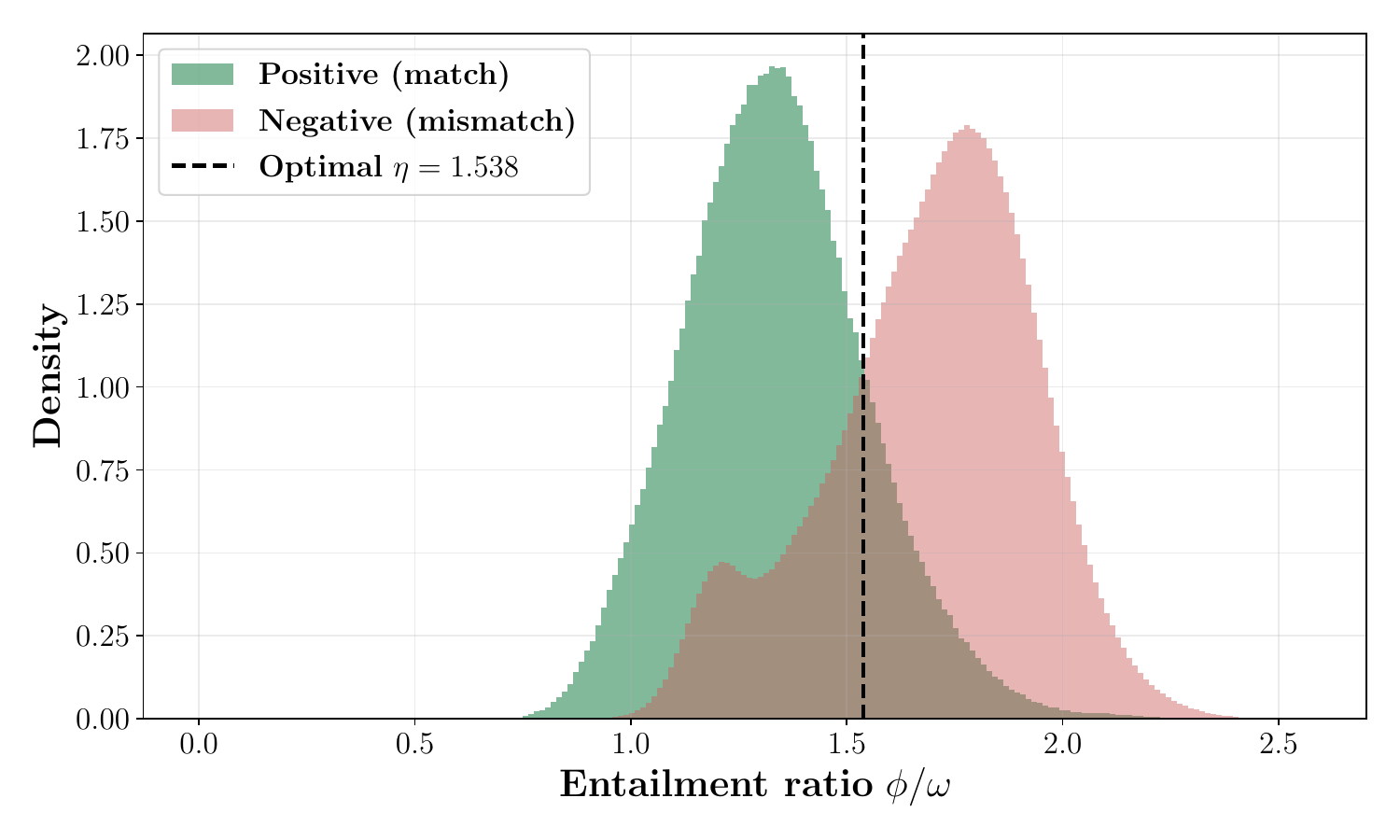}
        \caption{ImageNet}
        \label{fig:imagenet}
    \end{subfigure}
    
    \caption{Image-text entailment distributions. The plot shows the entailment ratio $\phi/\omega$, or the minimum $\eta_{img}$ required for entailment. The dotted line shows Youden's J statistic, or the optimal $\eta_{img}$ that separates the positive and negative distributions. The overlap in the two distributions shows that false positives and negatives are intrinsic to the backbone's pre-trained geometry, which we mitigate by selecting the threshold $\eta_{img}$ that maximizes separation. }
    \label{fig:combined_image}
\end{figure}

This statistic captures the optimal trade-off point where the model maximizes the recovery of true semantic hierarchies while minimizing spurious activations from unrelated concepts. 

We find the optimal $\eta_{img}$ values to be $1.372$ for CIFAR100, and $\eta_{img}=1.538$ for Imagenet. We round these to $1.4$ and $1.5$ in our experiments.
\begin{figure}[h]
    \centering
    \includegraphics[width=0.7\linewidth]{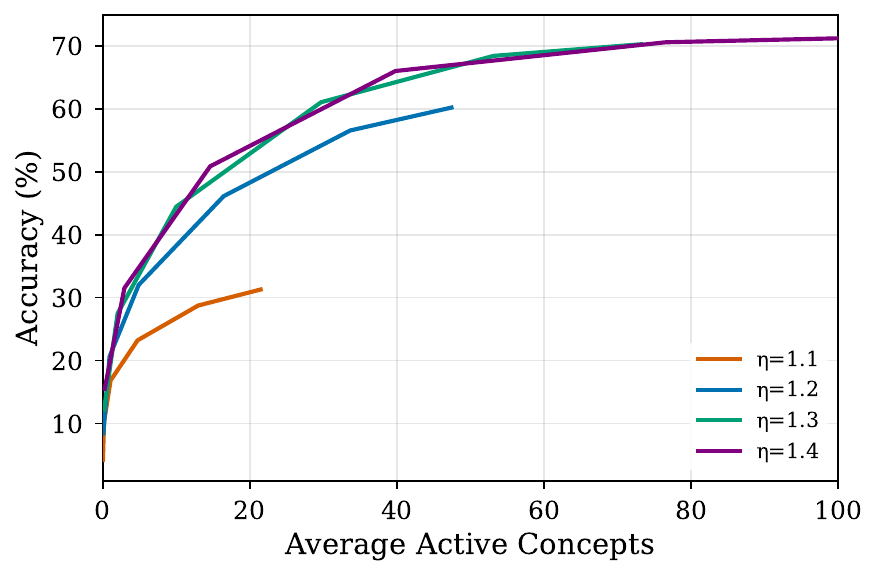}
    \caption{\textbf{Accuracy vs. number of active concepts, SUN397.} We sweep $\eta_{img}$ on a validation set to determine the optimal value on datasets where the distribution shift between proxy class labels and the concept bank is large.}
    \label{fig:sunstability}
\end{figure}
\paragraph{SUN397 \& CUB-200} 
For the SUN397 and CUB-200 datasets, this method of selecting the optimal $\eta_{img}$ resulted in poor performance, likely due to a larger distribution shift between the proxy class labels and the used concept bank. As an alternative option for datasets like these, we tune $\eta_{img}$ by sweeping a range of candidate values, and analyzing the performance on a validation split of the training set. We select $\eta_{img}=1.4$ and $\eta_{img}=1.5$ for SUN397 and CUB-200 respectively. See Fig.~\ref{fig:sunstability} for an example on SUN397.

\subsection{Finding the Optimal Intra-Modal $\eta_{text}$}
\label{app:eta_text}
\paragraph{Experimental Setup.}
Due to the exponential expansion of volume in hyperbolic space, the aperture required to capture a semantic subtree varies drastically with hierarchy depth. A fixed $\eta_{text}$ that is sufficient for general concepts (low norm) is often too narrow for specific concepts (high norm), leading to false negatives in entailment detection.

To derive a robust scaling law for concept-to-concept entailment, we performed a controlled experiment using WordNet hierarchies as a ground-truth reference. We initiate a crawler from 15 diverse mid-level roots (e.g., \textit{mammal}, \textit{tool}, \textit{structure}) to generate candidate parent-child pairs, across 3 depths.

\paragraph{Evaluation metrics.}
To ensure the derived statistics were applicable to our trained model, we applied a few filters to the candidate pairs:
\begin{itemize}
    \item \textbf{Norm Consistency}: We kept only pairs where both parent and child embeddings had a norm of $\|\mathbf{\tilde{c}}\|>\|\mathbf{\tilde{c}}_{min}\|$, in our case 0.27.
    \item \textbf{Hierarchical Integrity}: Secondly, we filtered for pairs satisfying $\|\mathbf{\tilde{c}}_{parent}\|<\|\mathbf{\tilde{c}}_{child}\|$ ensuring the parent is closer to the origin than its descendant.
\end{itemize}

For each of the $N=4504$ valid pairs, we compute the required entailment ratio ($\eta_{text}$), defined as the scalar multiplier required for the parent's cone to strictly contain the child: 
\begin{equation} \eta_{\text{text}} = \frac{\phi(\mathbf{c}_{\text{child}},\mathbf{c}_{\text{parent}})}{\omega(\mathbf{c}_{\text{parent}})} \end{equation} 
where $\phi$ is the exterior angle, and $\omega$ is the half-aperture.

\paragraph{Empirical Scaling Law.} 
Figure \ref{fig:scaling_law} in the main text illustrates the relationship between concept specificity (proxied by the hyperbolic norm $\|\mathbf{c}\|$) and the required threshold $\eta_{text}$. We observe a positive correlation ($r=0.729$), confirming that deeper concepts require wider entailment cones relative to their base aperture.

We fit a linear regression model to this distribution, yielding the relationship: \begin{equation} \eta_{text}(\|\mathbf{\tilde{c}}\|) \approx 8.62 \cdot (\|\mathbf{\tilde{c}}\| - 0.09). \end{equation}

\section{Cone parameterization and Distribution of Concept and Image Embeddings}
\label{app:distribution}
Our half-aperture definition:
\[
\omega(\mathbf{c}_i) = \sin^{-1}\left(\frac{2K}{\sqrt{c}\,\|\mathbf{\tilde{c}_i}\|}\right)
\]
depends inversely on the spatial concept embedding norm $\|\mathbf{\tilde{c}_i}\|$. This formulation renders the aperture sensitive to the scale of the embeddings: if the argument $u_i = \frac{2K}{\sqrt{c}\,\|\mathbf{\tilde{c}_i}\|}$ exceeds $1$, the cone angle becomes undefined (or, effectively, fully saturated at $\pi/2$). To ensure semantically meaningful cones, we carefully calibrate the scaling factor $K$ against the empirical norm distribution of the concept bank.

\begin{figure}[h]
    \centering
    % --- Subfigure A: Aperture Sensitivity (Bigger, 58% width) ---
    \begin{subfigure}[b]{0.55\linewidth}
        \centering
        \includegraphics[width=\linewidth]{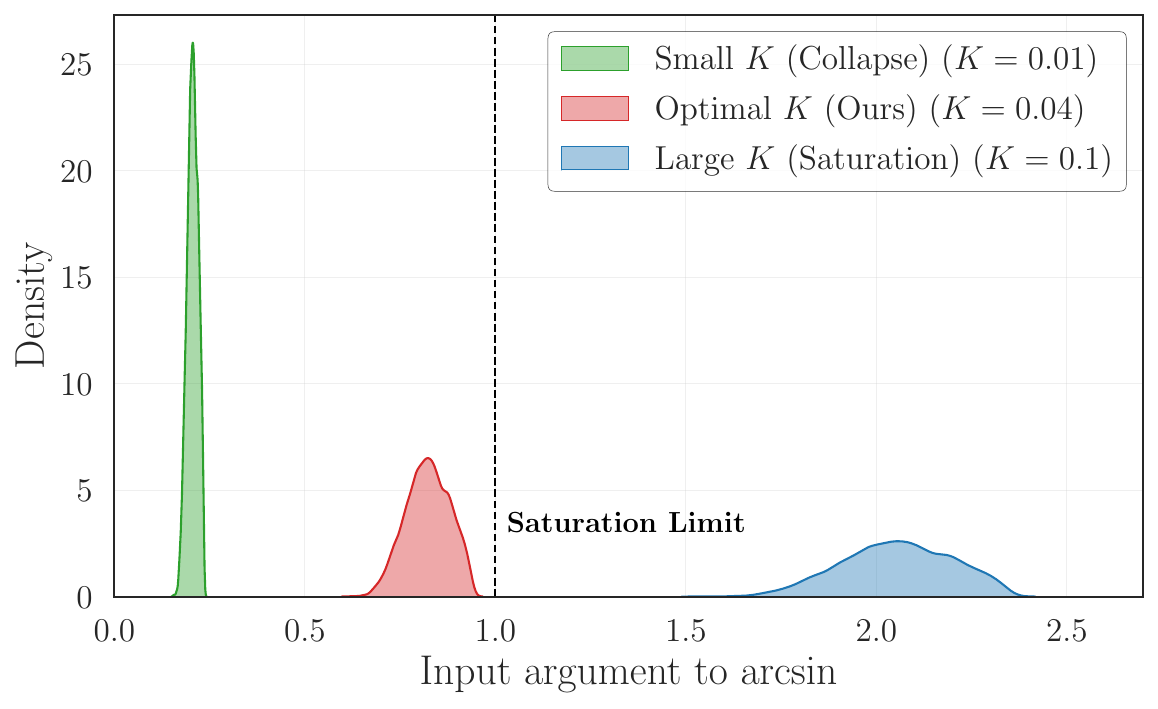}
        \caption{\textbf{Aperture Calibration.}}
        \label{fig:aperture_calibration}
    \end{subfigure}
    \hfill % Adds spacing between the two plots
    % --- Subfigure B: Norm Distribution (Smaller, 38% width) ---
    \begin{subfigure}[b]{0.44\linewidth}
        \centering
        \includegraphics[width=\linewidth]{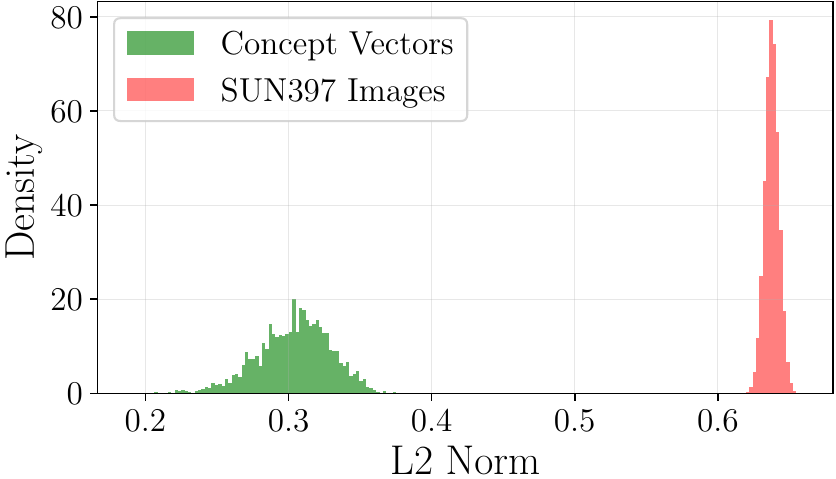}
        \vspace{1em}
        \caption{\textbf{Modality Gap.}}
        \label{fig:norm_dist}
    \end{subfigure}
    
    \caption{\textbf{Geometric Properties and Calibration.} \textbf{(a)} We calibrate the cone scaling factor $K$ to ensure that the distribution of concept apertures is well-posed (i.e. all arguments to $\arcsin$ are smaller than 1), avoiding numerical saturation limits (dashed line). For the default value of $K=0.1$, we observe that all possible text embeddings are clipped to 1, leading to a constant half-aperture of $\omega(\mathbf{c_i})=\tfrac{1}{2}\pi$, $\forall i$. \textbf{(b)} We visualize the native norm distribution of the backbone's representations, which informs our choice of filtering thresholds and scaling parameters. This plot highlights the \textit{modality gap}, where representations for the text and image modalities are clearly separated by norm.}
    \label{fig:geometric_calibration}
\end{figure}

\paragraph{Calibration and Saturation.}
As shown in Figure~\ref{fig:norm_dist}, the backbone exhibits a distinct \textit{modality gap}, where text-derived concept embeddings possess significantly smaller norms than image embeddings. This low-norm regime puts concepts at risk of saturation.
Figure~\ref{fig:aperture_calibration} illustrates this sensitivity. With a standard scaling of $K=0.1$ (blue), the argument $u_i$ exceeds the saturation limit ($1.0$) for the entire concept bank. This forces the half-aperture to clip to $\pi/2$, effectively treating all concepts as "universal" and destroying the hierarchical granularity of the bottleneck. Conversely, a very small $K=0.01$ (green dotted line) collapses the distribution near zero, resulting in "needle" cones with no entailment capacity.

We select $K=0.04$ (red) to balance this trade-off. For our filtered concept bank (where we enforce $\|\mathbf{\tilde{c}_i}\| \geq 0.27$), this choice ensures that the aperture arguments are widely distributed across the valid range $[0, 1)$, maximizing geometric expressivity. Consequently, the argument to the arcsine function is maximized at the lower bound of our filter, $\|\mathbf{\tilde{c}_i}\|=0.27$, which yields:
\[
u_{\max} \approx 0.937
\]
staying safely below the singularity at $1.0$. As a final safeguard for numerical stability, our implementation clamps the argument prior to the $\arcsin$ operation:
\[
\tilde{u}_i = \mathrm{clip}\left(u_i, -1 + \epsilon,\, 1 - \epsilon\right), \quad \epsilon=10^{-6}
\]
ensuring that $\omega(\mathbf{\tilde{c}_i})$ remains well-defined regardless of any potential outlier norms or floating-point variations. This clamping logic also applies for the input to the $\arccos$ for the definition of the exterior angle (Eq.~\ref{eq:exterior}). Together, the analytic constraint on $K$ and the norm-based filtering step ensure that the cone parameterization is numerically stable and semantically diverse.

\section{Qualitative Examples}
\label{qualitative}
\begin{figure}[h]
    \centering
    % First Subfigure (Hunting Lodge)
    \begin{subfigure}[b]{0.49\linewidth}
        \centering
        \includegraphics[width=\linewidth]{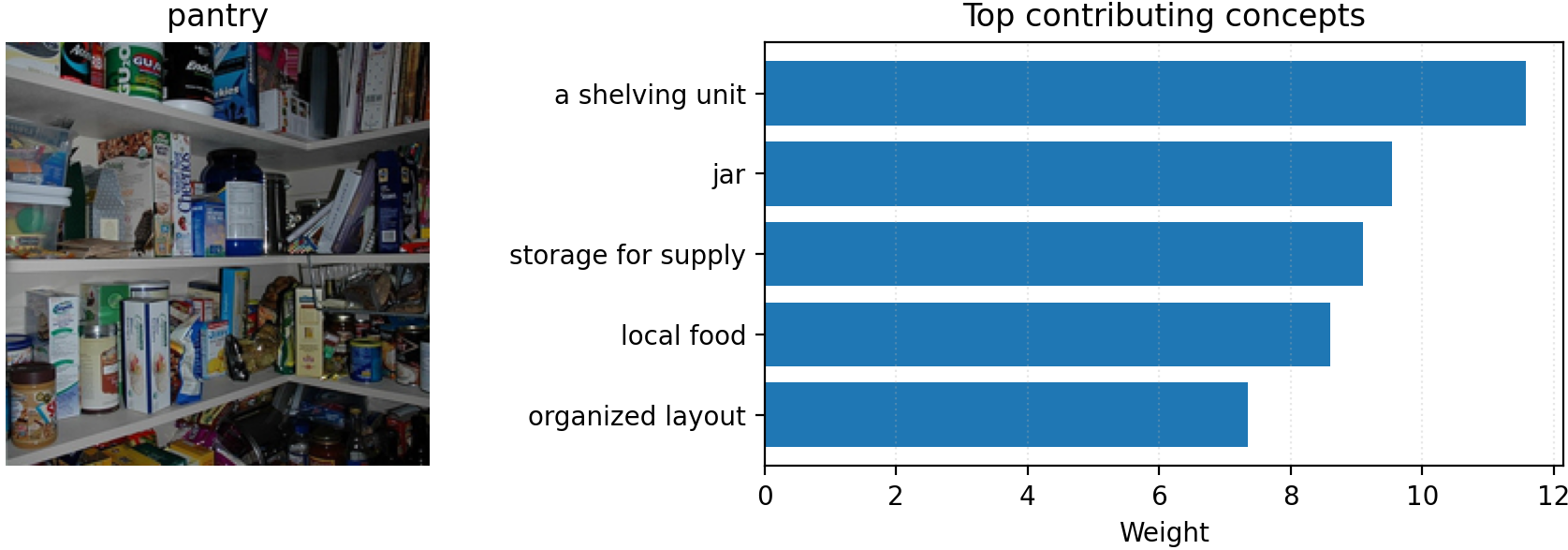}
        \caption{Pantry}
        \label{fig:lodge}
    \end{subfigure}
    \hfill % Adds spacing between the two images
    % Second Subfigure (Ocean)
    \begin{subfigure}[b]{0.49\linewidth}
        \centering
        \includegraphics[width=\linewidth]{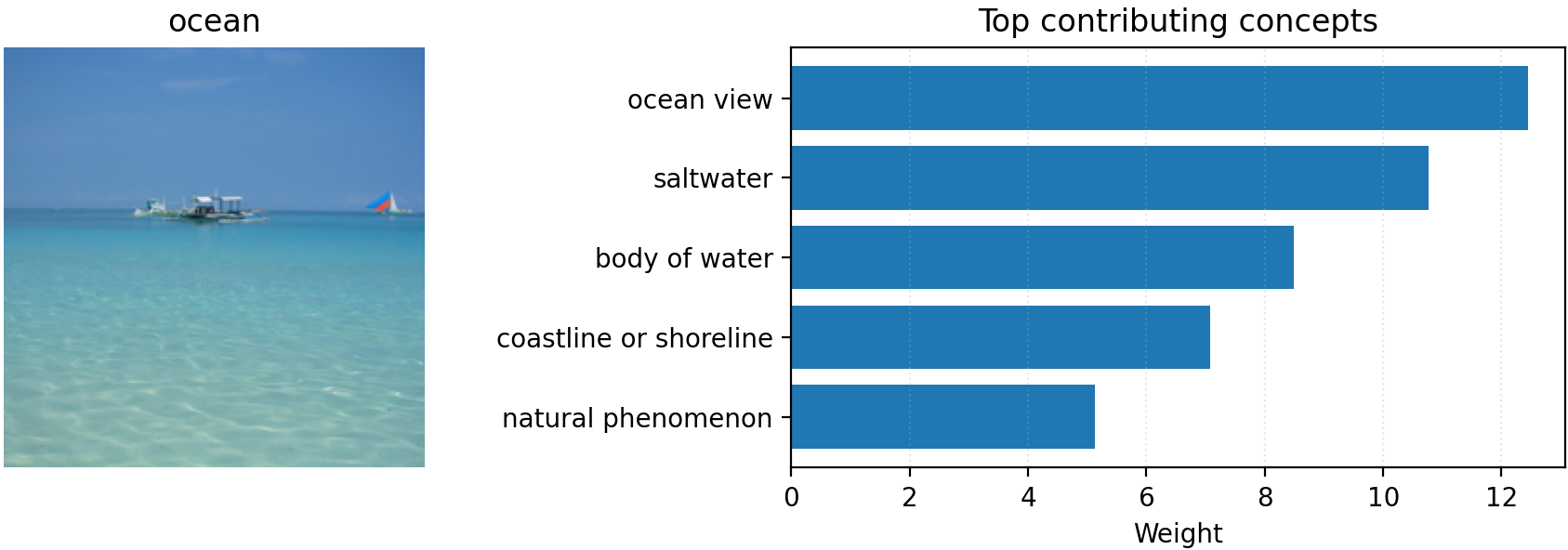}
        \caption{Ocean}
        \label{fig:ocean}
    \end{subfigure}
    
    \caption{\textbf{Global Explanations.} We visualize the top contributing  concepts (weight $\times$ activation) for the classes 'pantry' and 'ocean', along with two sample images from SUN397. More examples comparing LF-CBM and HypCBM are in the supplementary material.}
\end{figure}

\begin{figure}[h]
        \centering
        \includegraphics[width=\linewidth]{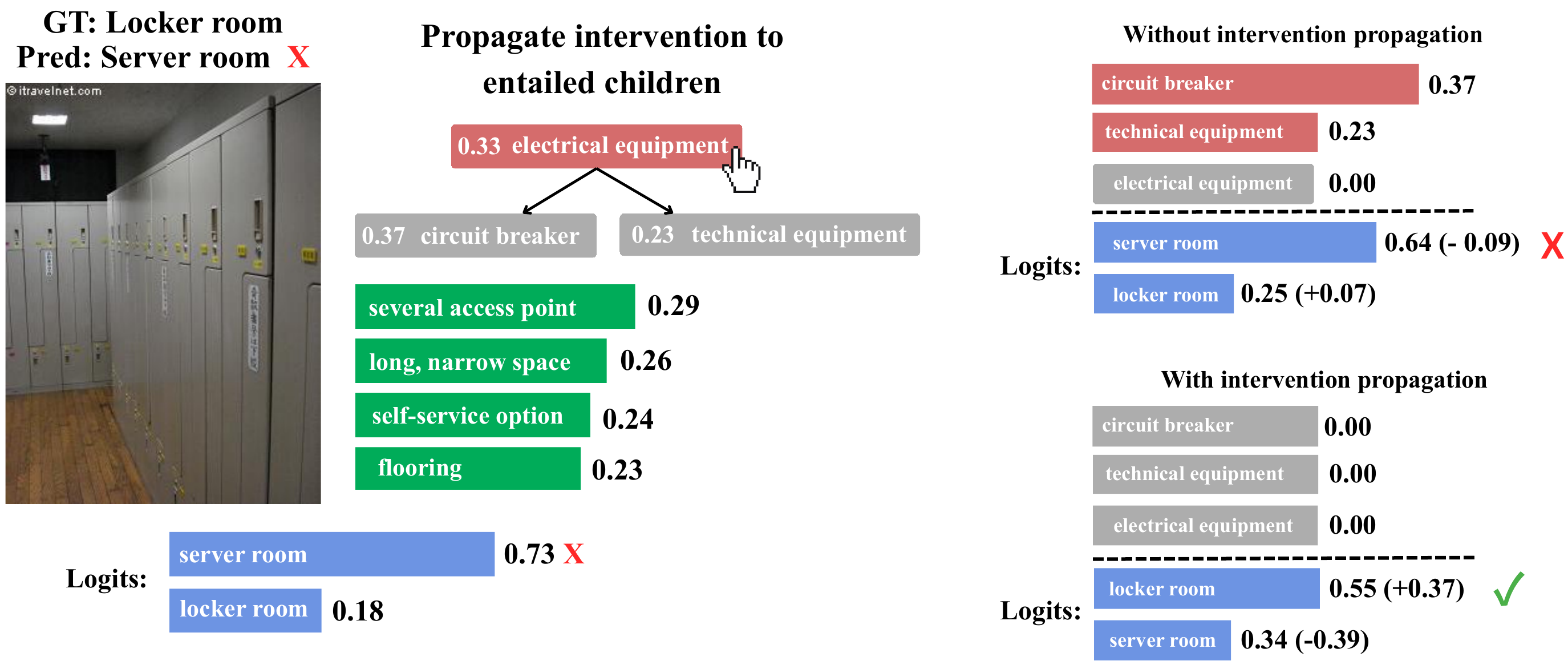}
        \caption{\textbf{Intervention Propagation.} An example that shows our intervention propagation mechanism on an image of a locker room that is misclassified as 'server room'. When intervening on 'electrical equipment', HypCBM automatically intervenes on entailed children 'circuit breaker' and 'technical equipment' too. Without this propagation, the prediction is still wrong, whereas with propagation the prediction flips to the correct class, 'locker room'.}
        \label{fig:qualitative_activations}
\end{figure}

% \begin{figure}[t]
%     \centering
%     % Top Panel: Original
%     \begin{subfigure}[b]{\linewidth}
%         \centering
%         \includegraphics[width=\linewidth]{figures/pantry.png}
%         \caption{\textbf{Original Prediction:} The model misclassifies a pantry as a living room.}
%         \label{fig:pantry_original}
%     \end{subfigure}
    
%     \vspace{0.5em} % Adds a little breathing room between images

%     % Bottom Panel: Intervened
%     \begin{subfigure}[b]{\linewidth}
%         \centering
%         \includegraphics[width=\linewidth]{figures/pantry_intervened.png}
%         \caption{\textbf{After Intervention:} Suppressing the parent concept 'furniture modifies the downstream predictions and correctly flips the prediction correctly to 'pantry'.}
%         \label{fig:pantry_intervened}
%     \end{subfigure}

%     \caption{\textbf{Intervention Example.} (a) The initial concept activations and prediction. (b) By intervening on the bottleneck (e.g., suppressing a specific concept), we observe a consistent shift in both the active child concepts and the final class prediction, demonstrating the model's responsiveness to human correction.}
%     \label{fig:intervention_qualitative}
% \end{figure}
\FloatBarrier
\section{Implementation Details}
\label{app:implementation}
\subsection{ANEC Interpolation and Pareto Frontier Estimation} 
\label{app:interpolation}
To accurately estimate the Accuracy at Number of Effective Concepts (ANEC) for specific integer budgets K, we employ piecewise linear interpolation between measured operating points. To ensure the fidelity of this approximation, we utilize an adaptive grid refinement strategy. We first perform a coarse sweep of regularization strengths $\lambda \in [10^{-1}, 10^{-7}]$ on a logarithmic scale. We then iteratively refine the grid by sampling additional $\lambda$ values in regions where the change in the number of active concepts is large, ensuring that the accuracy-sparsity curve is densely sampled near the target budgets (e.g., K=5,10,20).

\subsection{Training Time}
The computation of the entailment activation matrix scales linearly with the number of samples $N$, the concept bottleneck size $M$, and the embedding dimension $D$ ($D=512$ for HyCoCLIP), resulting in a time complexity of $\mathcal{O}(NMD)$. Specifically, the pairwise Lorentz exterior angle $\phi(\mathbf{z}, \mathbf{c})$ relies on the Minkowski dot product, which requires the same number of floating-point operations (FLOPs) as the standard Euclidean dot product used in linear classifiers. 

To maintain memory efficiency, we use batch size $B=512$, ensuring that the peak GPU memory usage scales as $\mathcal{O}(BM)$ rather than $\mathcal{O}(NM)$. This allows HypCBM to scale to thousands of concepts (e.g., ImageNet, $M=2950$) on consumer-grade hardware. Consequently, the primary computational bottleneck is not the geometric activation, but the optimization of the sparse linear classifier. Training time for CIFAR100 is approximately 2-3 minutes, whereas this scales from $\approx30$ minutes on SUN397 to $\approx5$ hours for ImageNet, depending on the regularization strength, $\lambda$. For very low values of $\lambda$ ($\lambda<1e-7$), the linear solver takes significantly longer.

Due to the numerical sensitivity of hyperbolic operations in the Lorentz model, we perform all geometric computations involving hyperbolic embeddings in double precision (FP64).

\section{Comprehensive CBM Taxonomy}
\label{app:taxonomy}
\begin{table}[h]
\centering
\caption{\textbf{Comprehensive taxonomy of CBM methodologies.} While recent methods successfully enforce bottleneck structure, they rely on task-specific training, graph priors, or auxiliary supervision (e.g., bounding boxes or logical rules). HypCBM is the only framework that achieves a structured bottleneck completely post-hoc, requiring no learned modules or auxiliary supervision beyond the pre-trained manifold.}
\resizebox{\textwidth}{!}{
\begin{tabular}{lcccccl}
\toprule
\textbf{Method} & \textbf{Post-hoc} & \textbf{Zero-shot} & \textbf{Structured} & \textbf{Backbone} & \textbf{Auxiliary Supervision} & \textbf{Learned Modules}\\
\midrule
LaBo \cite{yang2023languagebottlelanguagemodel} & \cmark & \cmark & \xmark & Any VLM & Concept Optimization & Optimization-based selection  \\
CEM \cite{zarlenga2022conceptembeddingmodelsaccuracyexplainability} & \xmark & \xmark & \xmark & Task-specific & Concept Labels & Soft concept embeddings \\
VLG-CBM \cite{srivastava2025vlgcbmtrainingconceptbottleneck} & \xmark & \xmark & \xmark & Any VLM & Object Bounding Boxes & Learned CBL projection \\
DN-CBM \cite{rao2024discoverthennametaskagnosticconceptbottlenecks} & \xmark & \xmark & \xmark & CLIP \cite{radford2021learning} & \textbf{None} & Sparse Autoencoder \\
Hierarchical CBM \cite{pittino2023hierarchical} & \xmark & \xmark & \cmark & Task-specific & Hierarchy Labels & Supervised hierarchy layers \\
C2F-CBM \cite{panousis2024coarse} & \xmark & \xmark & \cmark & Task-specific & Concept Labels & Probabilistic modules  \\
LogicCBM \cite{vemuri2025logiccbmslogicenhancedconceptbasedlearning} & \xmark & \xmark & \cmark & Task-specific & Logical Rules & Neuro-symbolic reasoning \\
Relational CBM \cite{barbiero2024relational} & \xmark & \xmark & \cmark & Task-specific & Relational Graphs & Relational graph/weights \\
Graph CBM \cite{xu2026graph}/CRM \cite{debot2025interpretablehierarchicalconceptreasoning} & \xmark & \xmark & \cmark & Task-specific & Relational Graphs & Graph reasoning \\
PCBM \cite{yuksekgonul2023posthoc} / LF-CBM \cite{oikarinen2023labelfree} & \cmark & \cmark & \xmark & Any VLM & \textbf{None} & \textbf{None} \\
\midrule
\textbf{HypCBM (Ours)} & \cmark & \cmark & \cmark & \textbf{Hyperbolic VLM} & \textbf{None} & \textbf{None} \\
\bottomrule
\end{tabular}}
\end{table}

\end{document}